\documentclass[letterpaper]{article} 
\usepackage{aaai24}  
\usepackage{times}  
\usepackage{helvet}  
\usepackage{courier}  
\usepackage[hyphens]{url}  
\usepackage{graphicx} 
\usepackage{natbib}  
\usepackage{caption} 
\usepackage{subcaption}
\usepackage{algorithm}
\usepackage{algorithmic}
\newcommand{\algorithmicbreak}{\textbf{break}}
\newcommand{\BREAK}{\STATE \algorithmicbreak}

\usepackage{newfloat}
\usepackage{listings}
\DeclareCaptionStyle{ruled}{labelfont=normalfont,labelsep=colon,strut=off} 
\floatstyle{ruled}
\newfloat{listing}{tb}{lst}{}
\floatname{listing}{Listing}
\usepackage{makecell}

\usepackage{amsmath, amsfonts} 
\usepackage{amssymb}

\DeclareMathOperator*{\argmin}{arg\,min}
\DeclareMathOperator{\Var}{\mathrm{V}}
\DeclareMathOperator{\E}{\mathbb{E}}
\DeclareMathOperator{\V}{\mathbb{V}}
\newcommand{\U}{\mathrm{U}}
\newcommand{\CV}{\mathrm{CV}}
\DeclareMathOperator{\diag}{diag}

\usepackage{physics}

\usepackage{centernot}
\usepackage{mathtools}
\usepackage{stmaryrd}
\makeatletter
\newcommand{\xMapsto}[2][]{\ext@arrow 0599{\Mapstofill@}{#1}{#2}}
\def\Mapstofill@{\arrowfill@{\Mapstochar\Relbar}\Relbar\Rightarrow}
\makeatother

\title{Adaptive Uncertainty-Guided Model Selection for Data-Driven PDE Discovery}
\author {
    Pongpisit Thanasutives\textsuperscript{\rm 1},
    Takashi Morita\textsuperscript{\rm 2},
    Masayuki Numao\textsuperscript{\rm 2},
    Ken-ichi Fukui\textsuperscript{\rm 2}
}
\affiliations {
    \textsuperscript{\rm 1}Graduate School of Information Science and Technology, Osaka University, Osaka, Japan\\
    \textsuperscript{\rm 2}SANKEN (The Institute of Scientific and Industrial Research), Osaka University, Osaka, Japan\\
    \{thanasutives, t-morita, numao, fukui\}@ai.sanken.osaka-u.ac.jp
}
\begin{document}
\maketitle
\begin{abstract}
We propose a new parameter-adaptive uncertainty-penalized Bayesian information criterion (UBIC) to prioritize the parsimonious partial differential equation (PDE) that sufficiently governs noisy spatial-temporal observed data with few reliable terms. Since the naive use of the BIC for model selection has been known to yield an undesirable overfitted PDE, the UBIC penalizes the found PDE not only by its complexity but also the quantified uncertainty, derived from the model supports' coefficient of variation in a probabilistic view. We also introduce physics-informed neural network learning as a simulation-based approach to further validate the selected PDE flexibly against the other discovered PDE. Numerical results affirm the successful application of the UBIC in identifying the true governing PDE. Additionally, we reveal an interesting effect of denoising the observed data on improving the trade-off between the BIC score and model complexity. Code is available at \url{https://github.com/Pongpisit-Thanasutives/UBIC}.
\end{abstract}
\section{Introduction}
Data-driven discovery has emerged as a popular approach for uncovering the governing partial differential equation (PDE) of a dynamical system, offering flexibility and satisfactory accuracy without relying heavily on domain knowledge. Sparse regression is typically leveraged to approximate a linear combination of user-specified candidate terms that optimally balances between a capability to predict temporal dynamics and the model complexity, hence so-called SINDy (sparse identification of nonlinear dynamics)-based approaches \cite{brunton2016discovering}, which were successfully applied in diverse disciplines from aerodynamics \cite{aerodynamics}, biological transport \cite{biology} to epidemiology \cite{epidemiology}. 

Diverse regularized regressors, designed to achieve the sparse identification were, for example, STRidge (sequential threshold ridge regression in PDE-FIND) \cite{rudy2017data}, LASSO (least absolute shrinkage and selection operator) \cite{LASSO} and SR3 (Sparse relaxed regularized regression) \cite{SR3}. However, tuning the regularization hyperparameter(s) of these methods for model selection is challenging, because the chosen hyperparameter may deliver an overfitted model. Also, it is difficult to control the model complexity. As a result, there is a risk of overlooking the true governing equation of a particular complexity. 

The mixed-integer optimization (MIO) for best-subset selection \cite{MIO} has been introduced to customize the desirable sparsity in the provably optimal MIO-SINDy \cite{bertsimas2023learning}. Various best-subset solvers are utilized to collect potential PDEs, from which one is automatically selected as the most suitable underlying PDE by an algorithm \cite{nPIML}. Although impressive progress has been made in deliberate consideration of likely PDEs consecutively arranged with a maximum complexity constraint, the important question remains: \textit{How do we select the best model that reveals the true governing PDE form?}. 

Before the model selection, we assure the existence of such a suitable PDE within our collection of potential models by enhancing the quality of the observed data, which are maybe noisy. A few previous attempts to denoise the distorted observed data (and its derivatives) were through derivative computation using polynomials, spline-based models, and Robust PCA (principal component analysis) \cite{rudy2017data, sun2022bayesian, li2020robust}. To simplify the experimental design, we concentrate on our preferred choices of denoising techniques that have numerically shown a positive impact on the model selection step: the regularized KSVD \cite{RKSVD}, a dictionary learning generalizing the K-means clustering. 

In the model selection given a finite set of potential PDEs, Akaike information criterion (AIC) \cite{AIC, mangan2017model, dong2023Aaic} and Bayesian information criterion (BIC) \cite{BIC} are commonly adopted as metrics evaluating point estimates of model parameters. However, in practice, the AIC and BIC values associated with a sequence of complexity-increasing linear models, fitted on an overcomplete candidate library to estimate temporal dynamics, tend to decrease, albeit insignificantly, after even the actual PDE is found. Therefore, naively selecting the PDE that minimizes the information criterion, could lead to erroneous equations \cite{nPIML}. 

\begin{figure}[t]
\centering
\includegraphics[width=\linewidth]{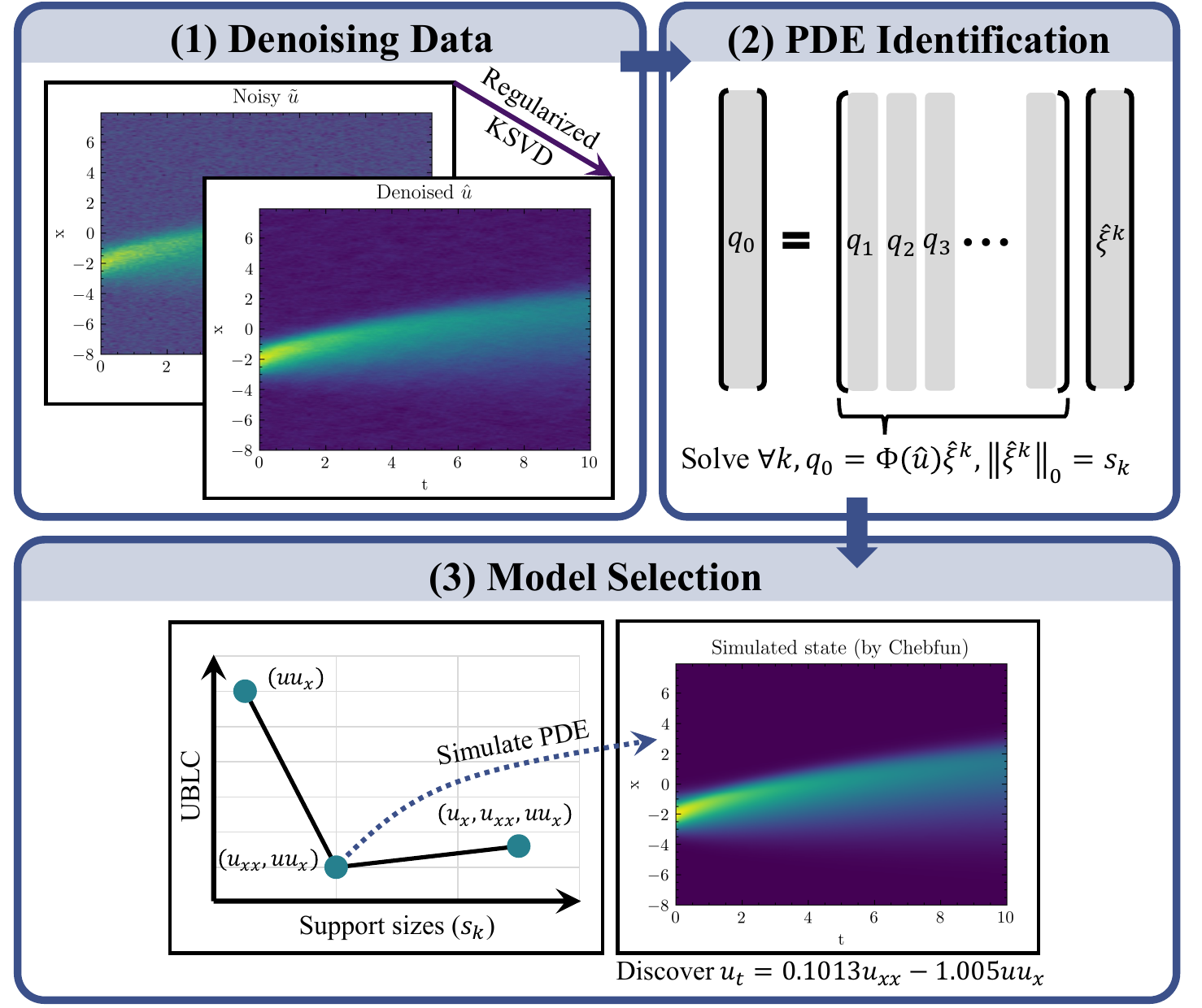}
\caption{Schematic diagram of the Burgers' PDE discovery example incorporating the UBIC for the model selection}
\label{fig:diagram}
\end{figure}

In this paper, we suggest a new way to assess the goodness of an approximated model by quantifying its associated uncertainty. We go beyond point estimates of the model parameters or coefficients and instead put a posterior belief on them. For each potential PDE, Bayesian linear regression is modeled on its corresponding subset to obtain the posterior distribution of the coefficients. Then, the coefficients' mean and covariance are computed, generally through certain posterior samples or analytical methods (if available), to quantity the uncertainty based on the coefficient of variation. By this means, we gain insights into the reliability of the estimated coefficients, which is, in turn, exploited adaptively by our proposed uncertainty-penalized Bayesian information criterion (UBIC). Figure \ref{fig:diagram} visualizes the primary steps to discover the underlying PDE starting from the denoising data to model selection. 

The UBIC-selected PDE can be optionally validated using simulation-based model selection, measuring the BIC of the PDE simulated state solution for directly predicting the denoised observed data. We investigate the use of a physics-informed neural network (PINN) \cite{PINN} as a differentiable automatic PDE solver. By incorporating physical laws as a part of its learning constraints, PINN enables the flexible PDE-solving approach. 

We list the main contributions in what follows. 
\begin{itemize}
    \item The UBIC integrates the quantified uncertainty of each potential PDE to penalize the BIC, resulting in the true identification of the parsimonious and stable governing PDE without heavy reliance on hyperparameter tuning. 
    \item We numerically exhibit the positive impact of denoising in terms of improving the trade-off given by the BIC. 
    \item We explore the simulation-based model selection that evaluates the efficiency of the PDE state solution, simulated by a PINN, in predicting the denoised data directly.
\end{itemize}

\subsubsection{Bayesian PDE Discovery Methods} Although such uncertainty-quantified SINDy (UQ-SINDy) \cite{UQ-SINDy} and threshold sparse Bayesian regression with error bars \cite{zhang2018robust}, were previously introduced, the uncertainty-guided model selection remains underexplored. Notably, the UQ-SINDy employed sparsifying priors to induce nonzeros terms identified with their posterior inclusion probabilities generally once, leading to risks of missing some correct terms or including incorrect ones. The latter work also risked excluding small yet important PDE coefficients due to a sensitive threshold. The comparisons are discussed thoroughly in the supplement. 
\section{Methodology}
\subsection{Problem Formulation}
Let us assume without loss of generality that the system state $u$ in a two-dimensional (2D) spatio-temporal grid of spatially distributed physical systems satisfies

\begin{equation}
    \partial_{t}u = \mathcal{N}(u, \partial_{x}u, \partial^{2}_{x}u, \dots; \xi).
\end{equation}

\noindent We aim to discover $\mathcal{N}$, a linear or nonlinear operator involving spatial derivatives of the state variable $u$ only. Parametric dependency $\xi_{j}$ is a constant vector-valued coefficient. For convenience, we consider $\partial_{x}u \equiv u_{x}$ and other notations alike. Since our observation input $\Tilde{u}$ may be disturbed with noise, or mathematically $\Tilde{u}_{ij} = u(x_{i}, t_{j}) + z(x_{i}, t_{j})$, we begin our PDE discovery approach by denoising on $\Tilde{u}$. 

Noise $z(x_{i}, t_{j}) \sim \frac{\epsilon\sigma_{u}}{100}\mathcal{N}(0, 1)$ is drawn from standard Gaussian distribution, and scaled proportionally to $\epsilon$\% of the standard deviation (sd) $\sigma_{u}$ calculated over the domain. 

\subsection{Denoising Data}
Suppose the grid is in a two-dimensional (2D) space, we turn $\Tilde{u}$ into a zero-mean array stacking flattened patches, regarded as signals $S_{p}(\Tilde{u}) \in \mathbb{R}^{p^{2} \times f}$; where $p$ and $f$ determine the patch size and the number of features. In Eq. (\ref{eq:rksvd}), we seek the dictionary $D$ of $c$ atoms and corresponding sparse code $A$ to represent $S_{p}(\Tilde{u})$, solving the following $\rho$-regularized dictionary learning problem: 

\begin{equation} \label{eq:rksvd}
\begin{aligned}
    \min_{D, A}&{{\norm{S_{p}(\Tilde{u})-DA}^{2}_{F}}+\rho\norm{A}^{2}_{F}}\\
    \text{subject to } &\norm{d_{j}}_{2} = 1,\, j = 1, \dots, c\\
    &\norm{a_{l}}_{0} \leq L,\, l = 1, \cdots, f.
\end{aligned}
\end{equation}

\noindent A couple of optimized $D$ and $A$ is obtain through regularized K-SVD training iterations. With the final fixed $D$, the ultimate sparse code $A$ is then found using the orthogonal matching pursuit (OMP) algorithm with $\lfloor\frac{p^{2}}{10}\rfloor$ transforming sparsity to reconstruct the denoised observed data $\hat{u} = S^{-1}_{p}(DA)$ from the patches, via the inverse $S^{-1}_{p}$. $\norm{\cdot}_{F}$ is the Frobenius matrix norm. We define $\mathcal{G}_{\hat{u}}(x_{i}, t_{j}) = \hat{u}_{ij}$ as the denoised state function. 

If we encounter 3D or 4D spatial-temporal data, denoised $\hat{u}$ is achieved efficiently by applying 2D Savitzky-Golay filters \cite{savgol_ref1, krumm2001savitzky}, which can be used with SVD (singular value decomposition). 

\subsection{PDE Identification}
Best-subset regression is used next to recover a sequence of potential parsimonious PDEs along their corresponding coefficients, with a maximal bound of support sizes (i.e., the number of nonzero coefficients). Basically, a careful forward-backward elimination algorithm \cite{guyon2002gene} can be implemented to approximate the best subsets. 

We compute a presumably overcomplete (with a maximum derivative order) library $\Phi(\hat{u})$, collecting candidate terms of the denoised observed data. According to the weak formulation \cite{wf}, $i$-th numerical value of $j$-th candidate (column-wise) in $\Phi(\hat{u}) \in \mathbb{R}^{N_{\Omega}\times N_{q}}$ is given by integrating over a local spatio-temporal subdomain $\Omega_{i}$, whose (rectangular) lengths are $H_{x}$ and $H_{t}$. 

\begin{equation} \label{eq:weak_form}
\begin{aligned}
    \Phi(\hat{u}) &= \mqty[\cdots & q_{j} & \cdots],\, j = 1, \dots, N_{q}; \\
    q^{i}_{j} &= \int_{\Omega_{i}} w\phi_{j} \,d\Omega,\, i = 1, \dots, N_{\Omega}.\\
\end{aligned}
\end{equation}

\noindent $\phi_{j}$ is regarded as a candidate function, for example, $\mathcal{G}^{2}_{\hat{u}}$ and $\partial^{2}_{x}\mathcal{G}_{\hat{u}}$. $N_{\Omega}$ is the number of domain centers; $\forall i,\, (x^{c}_{i}, t^{c}_{i})$. The smooth weight, e.g., $w = (\underline{x}^{2}-1)^{2}(\underline{t}^{2}-1)^{2};$ where $\underline{x} = (x-x^{c}_{i})/H_{x}, \underline{t} = (t-t^{c}_{i})/H_{t}$ conditioned by $(\underline{x}, \underline{t}) \in [-1, 1]^{2}$, is a viable function for discovering the Burgers' PDE as it vanishes along the boundary $\partial\Omega_{i}$. Certainly, higher polynomial orders are possible. By integration by parts on Eq. (\ref{eq:weak_form}), numerical noisy derivative evaluation of $\phi_{j}$ is supposed to be carried out on the noiseless $w$. We use the implementation provided in the PySINDy package \cite{pysindy_ref1, pysindy_ref2}. Remark that the noise-tolerant representation by the convolutional weak formulation (CWF) \cite{cwf} could be leveraged at the library construction stage as well. 

We attain an estimate $\hat{\xi}^{k}$ of the optimal PDE coefficients with $s_{k}$ support sizes by the best-subset selection:

\begin{equation} \label{eq:best_subset}
    \hat{\xi}^{k} = \argmin_{\xi^{k}}{\norm{q_{0}-\sum^{N_{q}}_{j=1}q_{j}\xi^{k}_{j}}^{2}_{2}}, \text{ subject to } \norm{\xi^{k}}_{0} = s_{k};
\end{equation}

\noindent where $q^{i}_{0} = \int_{\Omega_{i}} w\partial_{t}\mathcal{G}_{\hat{u}} \,d\Omega$, and $q_{0} \approx \sum^{N_{q}}_{j=1}q_{j}\xi^{k}_{j} = \Phi(\hat{u})\xi^{k}$. Best-subset solvers, we experiment with to yield potential PDEs for an increasing sequence of support sizes $(s_{k})^{N_{s}}_{k=1}$, are based on MIO, SOS-1-formulated (type-1 specially ordered sets) \cite{bertsimas2005optimization} MIO-SINDy, FROLS (forward regression with orthogonal least squares) \cite{frols_ref1, frols_ref2} and L0BnB (branch-and-bound framework for sparse regression) \cite{l0bnb}. 

\subsection{Uncertainty-penalized Bayesian Information Criterion (UBIC) for Adaptive Model Selection}
This section finds the best number of supports presented in Eq. (\ref{eq:best_subset}) within a given range. The base metric, on which we rely to penalize the maximized log-likelihood value of a regression model with its complexity, is the BIC: 

\begin{equation} \label{eq:BIC}
\begin{aligned}
    \textrm{BIC}(\hat{\xi}^{k}) &= -2\log L(\hat{\xi}^{k}) + \log(N_{\Omega})s_{k};\\
    \log L(\hat{\xi}^{k}) &= -\frac{N_{\Omega}}{2}\log(\frac{2\pi}{N_{\Omega}}\norm{q_{0}-\Phi(\hat{u})\hat{\xi}^{k}}^{2}_{2}).
\end{aligned}
\end{equation}

\noindent $L$ is the model likelihood. Assuming the true governing PDE is parameterized by the vector-valued coefficient that distributes with relatively less uncertainty, compared to those that build the other potential PDEs, we define the UBIC as

\begin{equation} \label{eq:UBIC}
\begin{aligned}
    \textrm{UBIC}(\xi^{k}, \lambda_{\U}) &= \textrm{BIC}(\xi^{k}_{\mu}) + \lambda_{\U}\log(N_{\Omega})\U^{k}\\
    &= -2\log L(\xi^{k}_{\mu}) + \log(N_{\Omega})(s_{k} + \lambda_{\U}\U^{k}), 
\end{aligned}
\end{equation}

\noindent where $\U^{k}$ represents an estimated total uncertainty for $\xi^{k}$, scaled proportionally to $\log(N_{\Omega})$, similar to the penalizing complexity in the BIC, for convenient unification. The data-dependent $\lambda_{\U}$ controlling influence of $\U^{k}$ on model selection is adaptively adjusted by Algorithm \ref{alg:1}. Typically, a lower UBIC means a better-discovered PDE. Bayesian linear regression probabilistic view is placed on $\xi^{k}$ with Gaussian conjugate prior $\mathcal{N}(\xi^{k} \mid \xi^{k}_{0}, \Var^{k}_{0})$ to setup the posterior: 

\begin{equation}
\begin{aligned}
    p(\xi^{k} \mid \Phi(\hat{u}), q_{0}, \sigma_{q}) &\sim \mathcal{N}(\xi^{k} \mid \xi^{k}_{0}, \Var^{k}_{0})\mathcal{N}(q_{0} \mid \Phi(\hat{u})\xi^{k}, \sigma^{2}_{q}\mathbf{I}_{N_{\Omega}})\\
    &= \mathcal{N}(\xi^{k} \mid \xi^{k}_{\mu}, \Var^{k});\\
    \xi^{k}_{\mu} &= \Var^{k}(\Var^{k}_{0})^{-1}\xi^{k}_{0} + \frac{1}{\sigma^{2}_{q}}\Var^{k}\Phi(\hat{u})^{T}q_{0},\\
     \Var^{k} &= \sigma^{2}_{q}(\sigma^{2}_{q}(\Var^{k}_{0})^{-1} + \Phi(\hat{u})^{T}\Phi(\hat{u}))^{-1}.
\end{aligned}
\end{equation}

\noindent Later presented experimental results are produced with $\xi^{k}_{0} = \hat{\xi}^{k}$ and $V^{k}_{0} = \mathbf{I}_{\norm{\hat{\xi}^{k}}_{0}}$ as an identity matrix of size $\norm{\hat{\xi}^{k}}_{0}$. Note that $\xi^{k}_{0} = \vec{0}$, reducing the posterior mean to ridge estimate, is also a feasible option. By maximum likelihood estimation (MLE), we accordingly set the error variance $\sigma^{2}_{q} = \E[(q_{0}-\Phi(\hat{u})\hat{\xi}^{k})^{2}]$. Inspired by the coefficient of variation formula, the uncertainty $\U^{k}$ is defined as 

\begin{equation}
    \U^{k} = \frac{\CV^{k}}{\min_{k}{\CV^{k}}};\, \CV^{k} = \frac{\norm{\diag(\Var^{k})^{\circ\frac{1}{2}}}_{1}}{\norm{\xi^{k}_{\mu}}_{1}} = \frac{\sum^{s_{k}}_{i=j}\sqrt{V^{k}_{ij}}}{\norm{\xi^{k}_{\mu}}_{1}}.
\end{equation}

\noindent Here we essentially compute the relative standard deviation of the covariance matrix $\Var^{k}$ by taking an element-wise square (the $\circ\frac{1}{2}$ exponent) root on its diagonal vector ($\diag$) and then the division by $\norm{\xi^{k}_{\mu}}_{1}$. Each $\CV^{k}$ is rescaled by its minimum respecting the range of available support sizes, resulting in $\U^{k}$ whose numerical value is comparable to $s_{k}$. 

Once $\U^{k}$ is obtained, we converge the UBIC by Algorithm \ref{alg:1}, iteratively decreasing $\lambda_{\U}$ from its maximum bound $\lambda^{\textrm{max}}_{\U}$ derived to maintain the influence of the log-likelihood value (presumably greater than zero) in Eq. (\ref{eq:UBIC}). We impose the constraint based on the discovered $s_{k}$-support PDE. 

\begin{equation} \label{eq:bound}
\begin{aligned}
    \log N_{\Omega}(s_{k}+\lambda_{\U}\U^{k}) &\leq \abs{-2\log\hat{L}(\xi^{k}_{\mu})};\, \lambda_{\U} \geq 0\\
    0 \leq \lambda_{\U} \leq \lambda^{\textrm{max}}_{\U} &= \max_{k}{\frac{1}{\U^{k}}(\frac{2\log\hat{L}(\xi^{k}_{\mu})}{\log N_{\Omega}}-s_{k})}
\end{aligned}
\end{equation}

\noindent Algorithm \ref{alg:1} finds a proper $\lambda_{\U} = 10^{\lambda}$ by reducing $\lambda$ iteratively. We track the current and competitive optimal support sizes $(s_{k^{*}}, s_{k^{c}})$, and test the terminate condition at line \ref{alg:terminate}, which essentially checks whether we have the increased complexity with unsatisfactory improvement (see $\tau$), or the decreased complexity with already satisfying improvement. Also, $\tau_{0}$ can be set adaptively, yet offering the same correct selection as the default value. Such an effective heuristic is $\tau_{0} = P_{75}(S)$; $S = \{\tau^{k^{2}}_{k^{1}} \mid k^{1}, k^{2} = \argmin(r)\,\, \textrm{s.t.}\,\, r = s_{k^{2}}-s_{k^{1}}>0,\,\, \textrm{and}\,\, \forall s_{k^{0}}<s_{k^{1}},\, \textrm{BIC}(\xi^{k^{2}}_{\mu}) < \textrm{BIC}(\xi^{k^{1}}_{\mu}) < \textrm{BIC}(\xi^{k^{0}}_{\mu})\}$, the $75^{\textrm{th}}$ percentile of successive improvement factors respecting just BIC-decreasing models. If an overfitted model is detected by line \ref{alg:warning}, we retry with a stricter percentile of $S$, e.g., $P_{80}(S)$. Contrarily, lessening $\tau_{0}$ helps discern selecting a supposedly underfitted $1$-support PDE. 

\begin{algorithm}[tb]
\caption{Find the optimal complexity $s_{k^{*}}$ by tuning $\lambda_{\U}$}
\textbf{Input:} $\Phi(\hat{u}), q_{0}$ and $\hat{\xi}^{k}$\\
\textbf{Parameter:} $\tau_{0}$: Improvement acceptance threshold (default $= 0.02$), $N_{\delta} = 3$: Number of evenly spaces in $[0, \log_{10}\lambda^{\textrm{max}}_{\U}]$\\
\textbf{Output}: The optimal support sizes $s^{*}_{k}$ and tuned UBIC's\\hyperparameter $\lambda_{\U}$
\begin{algorithmic}[1]
\STATE Compute $\forall k \leq N_{s},\, \Var^{k},\xi^{k}_{\mu}$ and $\U^{k}$,\\with Gaussian prior $\mathcal{N}(\xi^{k} \mid \hat{\xi}^{k}, \mathbf{I}_{\norm{\hat{\xi}^{k}}_{0}})$
\STATE Assign $\lambda \gets \log_{10}\max(\lambda^{\textrm{max}}_{\U}, 0)$ \COMMENT{$-\infty$ if $\lambda^{\textrm{max}}_{\U} < 0$}
\STATE Assign $\delta \gets \frac{\lambda}{N_{\delta}}$ and $\lambda^{c} \gets \lambda - \delta$ \COMMENT{next candidate for $\lambda$}
\STATE Compute $\forall k,\, \mathcal{I}_{k} \gets \textrm{UBIC}(\xi^{k}, 10^{\lambda})$ using $\xi^{k}_{\mu}$ and $\U^{k}$
\STATE Find $s_{k^{*}}$ where $k^{*} \gets \argmin_{k}{\mathcal{I}_{k}}$
\WHILE{$\lambda^{c} > 0$}
\STATE Compute $\forall k,\, \mathcal{I}^{c}_{k} \gets \textrm{UBIC}(\xi^{k}, 10^{\lambda^{c}})$
\STATE Find $s_{k^{c}}$ where $k^{c} \gets \argmin_{k}{\mathcal{I}^{c}_{k}}$
\STATE Assign $\Delta s \gets s_{k^{c}}-s_{k^{*}}$
\STATE Assign $\Delta\textrm{BIC} \gets \textrm{BIC}(\xi^{k^{c}}_{\mu})-\textrm{BIC}(\xi^{k^{*}}_{\mu})$
\STATE Assign $\tau \gets \tau^{k^{c}}_{k^{*}}$; $\tau^{k^{c}}_{k^{*}} = \abs{\Delta\textrm{BIC}/(\textrm{BIC}(\xi^{k^{*}}_{\mu})\Delta s)}$
\IF {$(\Delta s > 0$ but $(\Delta\textrm{BIC}>0$ or $\tau < \tau_{0}))$ or\\ $(\Delta s < 0$ but $\Delta\textrm{BIC}>0$ and $\tau > \tau_{0})$} \label{alg:terminate}
\BREAK \COMMENT{terminate condition detected}
\ENDIF
\STATE Assign $\lambda \gets \lambda^{c}$ and $\lambda^{c} \gets \lambda - \delta$
\STATE Assign $\forall k,\, \mathcal{I}_{k} \gets \mathcal{I}^{c}_{k}$ and $k^{*} \gets k^{c}$ \COMMENT{$s_{k^{*}} \gets s_{k^{c}}$}
\ENDWHILE
\IF {$\abs{\textrm{BIC}(\xi^{k^{*}}_{\mu})-\textrm{BIC}(\xi^{k^{*}-1}_{\mu})}/\abs{\textrm{BIC}(\xi^{k^{*}-1}_{\mu})} < \tau_{0}$} \label{alg:warning}
\STATE Consider an increased $\tau_{0}$ to prevent overfitting 
\ENDIF
\STATE \textbf{return} $s_{k^{*}}, \lambda_{\U}=10^{\lambda}$ and $\forall k,\, \mathcal{I}_{k}$ \COMMENT{used in plotting}
\end{algorithmic}
\label{alg:1}
\end{algorithm}

\subsection{Simulation-based Model Selection}
As Eq. (\ref{eq:best_subset}) optimizes for the regression model that fits $\Phi(\hat{u})$ to approximate $q_{0}$ (a weak form of $u_t$), the attained model do not offer the direct comparison metric to the PDE solution $u$. However, they are expected to yield the true PDE, reducing unnecessary costs of simulating false governing PDEs. 

As an aid validation process, which can be computationally expensive or numerically infeasible for some ill-posed PDEs, we solve the found PDEs most likely to be the true PDE form based on our UBIC scores. Well-known numerical PDE solvers based on spectral methods are available in Chebfun \cite{chebfun} and Dedalus \cite{dedalus}. The software can accurately solve canonical stiff PDEs given initial and boundary conditions. Nevertheless, when solving a PDE containing extraneous high-order derivatives, the simulated solution may explode over time. Mathematical analysis is needed to rigorously address ill-posed PDEs, characterized by numerical instability in their solutions. Though such analysis exceeds our scope, we give the flexible neural network based treatise. 

\subsubsection{Physics-informed Neural Network (PINN) Learning}
PINN learning is an alternative approach for solving PDEs. The learned solution not only satisfies a specified PDE but is also stabilized to fit observational data. The general principle is to learn the mapping function from spatio-temporal data (in $\mathcal{D} = \{(x^{\mathcal{D}}_{i}, t^{\mathcal{D}}_{j})\}$ split) to denoised $s_{k}$-support PDE solution: $\forall i,j:\, (x^{\mathcal{D}}_{i}, t^{\mathcal{D}}_{j}) \xmapsto{f_{\Theta_{k}}}{} \hat{u}^{\mathcal{D}}_{ij}$ by minimizing the physics-informed loss respecting the discovered function $\hat{\mathcal{N}}$. 

\begin{equation} \label{eq:pinn}
\begin{aligned}
    \Theta^{*}_{k} = &\argmin_{\Theta_{k}}\bigg(\norm{\mathcal{F}^{\mathcal{D}}_{\Theta_{k}}-\hat{u}^{\mathcal{D}}}^{2}_{F}+\\
    &\norm{\partial_{t}\mathcal{F}^{\mathcal{D}}_{\Theta_{k}}-\hat{\mathcal{N}}(\mathcal{F}^{\mathcal{D}}_{\Theta_{k}}, \partial_{x}\mathcal{F}^{\mathcal{D}}_{\Theta_{k}}, \partial^{2}_{x}\mathcal{F}^{\mathcal{D}}_{\Theta_{k}}, \dots; \hat{\xi}^{k})}^{2}_{F}\bigg)
\end{aligned}
\end{equation}

\noindent We collect $(\mathcal{F}^{\mathcal{D}}_{\Theta_{k}})_{ij} = f_{\Theta_{k}}(x^{\mathcal{D}}_{i}, t^{\mathcal{D}}_{j})$. $\mathcal{D}_{\textrm{Train}}$ is the train split containing subsampled discretized spatio-temporal points, on which the PINN is trained, and likewise $\mathcal{D}_{\textrm{Val}}$ for hold-out validation dataset. Automatic differentiation is used to compute derivative terms, e.g., $\partial_{x}\mathcal{F}^{\mathcal{D}}_{\Theta_{k}}$. $\Theta_{k}$ is optimized by the second-order LBFGS \cite{lbfgs}. Trainable $\hat{\xi}^{k}$ is initialized beforehand from Eq. (\ref{eq:best_subset}). 

Deciding whether the $s_{k}$-support PDE is better or worse than the $s_{k+1}$-support PDE, we adjust the BIC in Eq. (\ref{eq:BIC}) as the complexity penalization avoids choosing the overfitted PDE that also generate the PINN predicted solution close to $\hat{u}^{\mathcal{D}_{\textrm{Val}}}$ on the validation split. The state-level BIC reads

\begin{equation} \label{eq:pinn_bic}
\begin{aligned}
    \textrm{BIC}^{\hat{u}}_{\Theta^{*}_{k}}(\hat{\xi}^{k}) = &-\abs{\mathcal{D}_{\textrm{Val}}}\log(\frac{2\pi}{\abs{\mathcal{D}_{\textrm{Val}}}}\norm{\hat{u}^{\mathcal{D}_{\textrm{Val}}}-\mathcal{F}^{\mathcal{D}_{\textrm{Val}}}_{\Theta^{*}_{k}}}^{2}_{F})\\
    &+\log(\abs{\mathcal{D}_{\textrm{Val}}})(\abs{\Theta^{*}_{k}}+\norm{\hat{\xi}^{k}}_{0}); 
\end{aligned}
\end{equation}

\noindent where $\abs{\Theta^{*}_{k}}$ tells the network's number of trainable parameters. The modification involves utilizing the validation data points in $\mathcal{D}_{\textrm{Val}}$ for the comparison. If $\textrm{BIC}^{\hat{u}}_{\Theta^{*}_{k}}(\hat{\xi}^{k}) < \textrm{BIC}^{\hat{u}}_{\Theta^{*}_{k+1}}(\hat{\xi}^{k+1})$, it is advisable not to increase the support sizes to $s_{k+1}$ under a fair circumstance where the PINN architecture and learning procedure are identical. We adhere to the vanilla paradigm for simplicity, though the PINN training could be improved, e.g., using multi-task learning techniques \cite{thanasutives2021adversarial}, since the network may overfit on $\mathcal{D}_{\textrm{Train}}$, stuck in a physics-obeying local minimum \cite{Bajaj_2023}.
\begin{table*}[th]
\centering
\caption{PDE descriptions. The number of discretized spatial and temporal points are specified by the $(N_{x}, N_{y}, N_{z})$ and $N_{t}$.}
\setlength{\tabcolsep}{2pt}
\begin{tabular}{|c|c|c|c|c|}
\hline
Dataset & PDE & $N_{x}, N_{y}, N_{z}$ & $N_{t}$ & $\epsilon$ \\
\hline
Burgers & $\partial_{t}u = 0.1\partial^{2}_{x}u - u\partial_{x}u$ & $256$ on $[-8, 8]$ & $101$ on $[0, 10]$ & $30$ \\
KdV & $\partial_{t}u = -\partial^{3}_{x}u - u\partial_{x}u$ & $512$ on $[-20, 20]$ & $501$ on $[0, 40]$ & $30$ \\
KS & $\partial_{t}u = -\partial^{2}_{x}u -\partial^{4}_{x}u - u\partial_{x}u$ & $1024$ on $[0, 32\pi]$ & $251$ on $[0, 100]$ & $30$ \\
NS & $w_{t} = 0.01(w_{xx} + w_{yy}) - uw_{x} - vw_{y}$ & $325 \times 170$ on $[2, 8.48] \times [0.3, 3.68]$ & $151$ on $[0, 30]$ & $1$ \\
\hline
RD & \makecell[c]{$u_{t} = u-u^{3}+v^{3}-uv^{2}+u^{2}v+0.1(u_{xx}+u_{yy})$\\$v_{t} = v-u^{3}-v^{3}-uv^{2}-u^{2}v+0.1(v_{xx}+v_{yy})$} & $256 \times 256$ on $[-10, 10] \times [-10, 10]$ & $201$ on $[0, 10]$ & $10$ \\
\hline
GS & \makecell[c]{$u_{t} = 0.014-0.014u-uv^{2}+0.02(u_{xx}+u_{yy}+u_{zz})$\\$v_{t} = -0.067v+uv^{2}+0.01(v_{xx}+v_{yy}+v_{zz})$} & $128 \times 128 \times 128$ on $[-1.25, 1.25]^{3}$ & $100$ on $[0, 10]$ & $0.1$ \\
\hline
\end{tabular}
\label{tab:dataset}
\end{table*}

\begin{figure*}[t]
\begin{subfigure}{.245\textwidth}
  \centering
  \includegraphics[width=\linewidth]{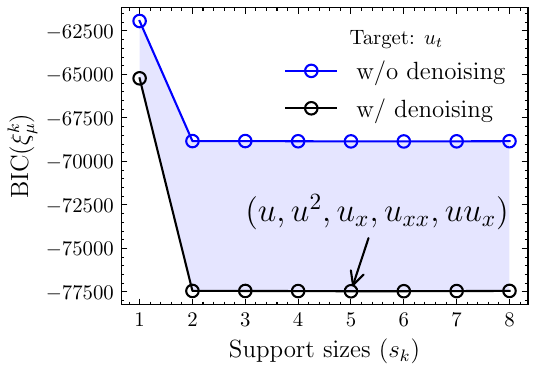}  
\end{subfigure}
\begin{subfigure}{.245\textwidth}
  \centering
  \includegraphics[width=\linewidth]{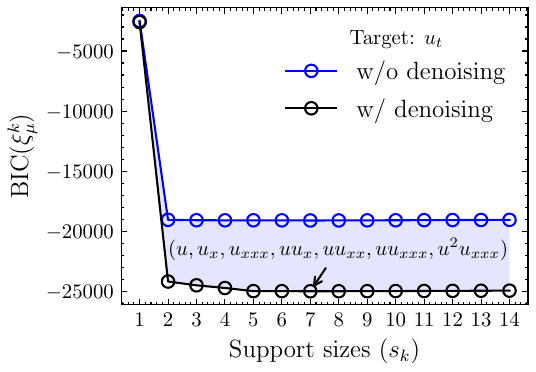}  
\end{subfigure}
\begin{subfigure}{.245\textwidth}
  \centering
  \includegraphics[width=\linewidth]{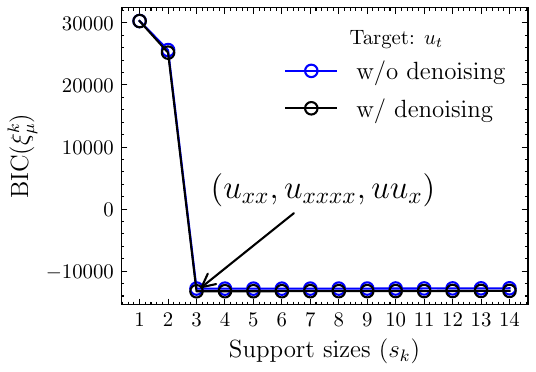}  
\end{subfigure}
\begin{subfigure}{.245\textwidth}
  \centering
  \includegraphics[width=\linewidth]{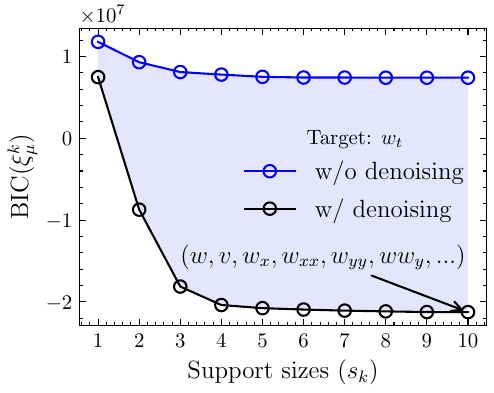}  
\end{subfigure}
\vfill
\begin{subfigure}{.245\textwidth}
  \centering
  \includegraphics[width=\linewidth]{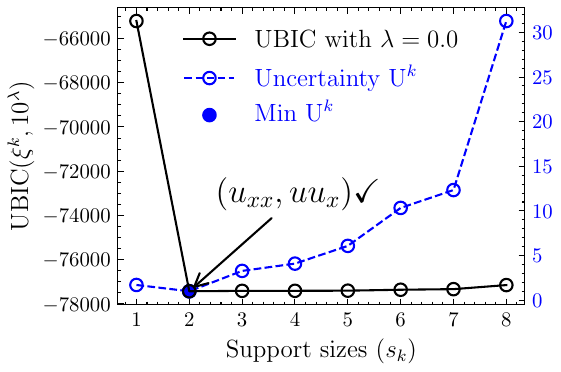}  
  \caption{Burgers}
  \label{fig:burgers}
\end{subfigure}
\begin{subfigure}{.245\textwidth}
  \centering
  \includegraphics[width=\linewidth]{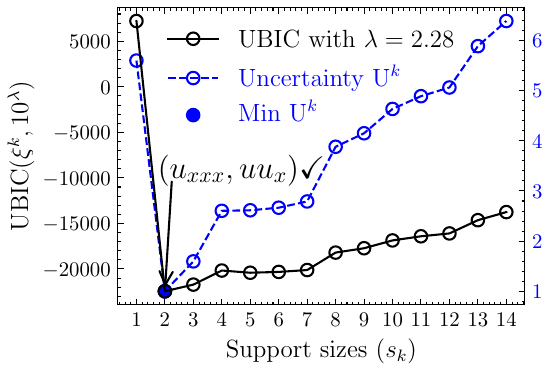}  
  \caption{KdV}
  \label{fig:kdv}
\end{subfigure}
\begin{subfigure}{.245\textwidth}
  \centering
  \includegraphics[width=\linewidth]{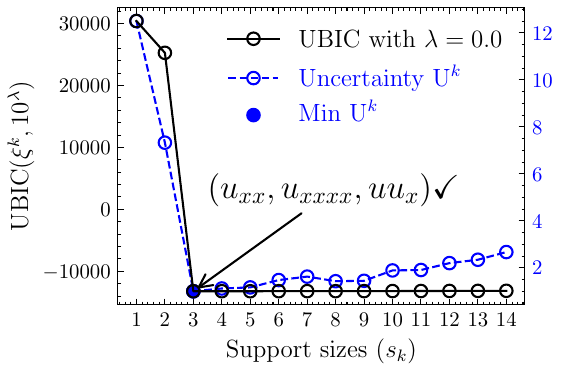}  
  \caption{KS}
  \label{fig:ks}
\end{subfigure}
\begin{subfigure}{.245\textwidth}
  \centering
  \includegraphics[width=\linewidth]{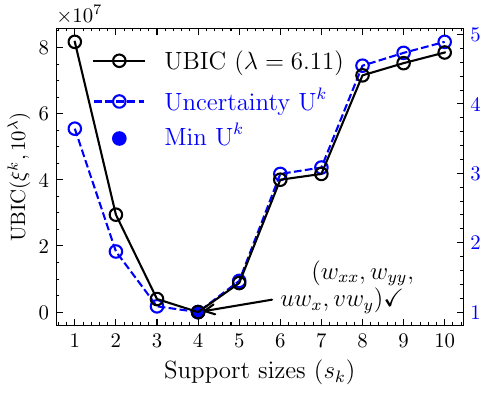}  
  \caption{NS}
  \label{fig:ns}
\end{subfigure}
\caption{We plot the BIC, uncertainty $\U^{k}$ and UBIC with tuned $\lambda_{\U} = 10^{\lambda}$ for the model selection in the Burgers, KdV, KS, and NS examples, arranged from left to right. ``$\checkmark$'' indicates that the UBIC selects the true PDE form.}
\label{fig:UBIC_res1}
\end{figure*}

\begin{figure*}[t]
\begin{subfigure}{.331\textwidth}
  \centering
  \includegraphics[width=\linewidth]{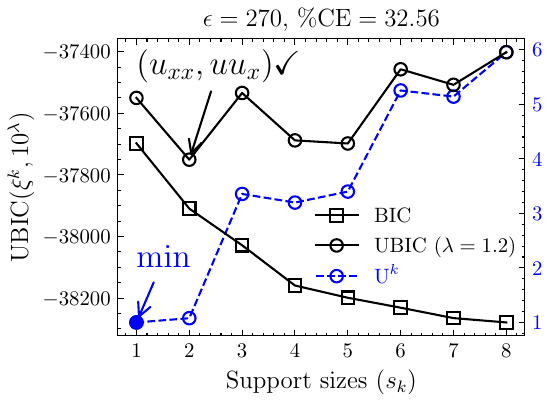}  
  \caption{Burgers}
\end{subfigure}
\begin{subfigure}{.331\textwidth}
  \centering
  \includegraphics[width=\linewidth]{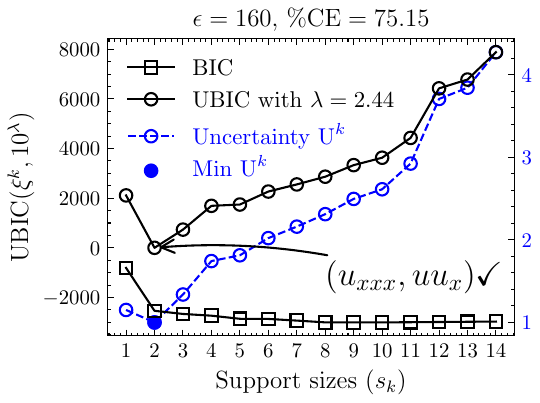}  
  \caption{KdV}
\end{subfigure}
\begin{subfigure}{.331\textwidth}
  \centering
  \includegraphics[width=\linewidth]{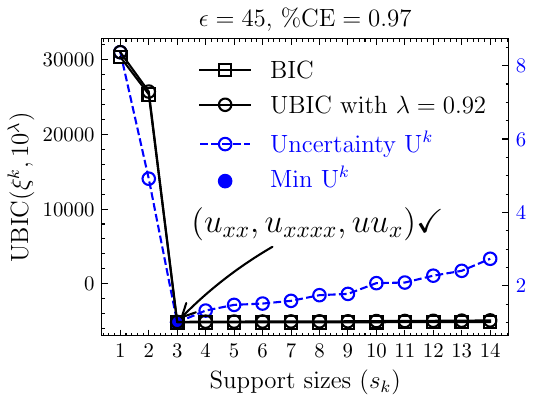}  
  \caption{KS}
\end{subfigure}
\caption{Robust adaptive model selection by the UBIC under the extremely noisy scenarios.}
\label{fig:extreme_noise}
\end{figure*}

\section{Numerical Result}
The PDE dataset description is given in Table \ref{tab:dataset}. Experiments were run on a $2.6$ GHz $6$-Core Intel i7 CPU with $32$ GB RAM. Further results and analysis are in the supplement. 

\subsubsection{Burgers' PDE} We tested the PDE solution with the initial condition: $u(x, 0) = e^{-(x+2)^{2}}$. To denoise $\Tilde{u}$, we ran regularized KSVD with $\rho = 0.05$ on the stack $S_{p}(\Tilde{u})$ created with square patches of size $8 \times 8$. For sparse encoding during the training, OMP was configured with one target sparsity. 

We gathered an overcomplete set of candidate functions $\phi_{j}(\cdot) \in \{(\mathcal{G}^{d_{1}}_{\hat{u}}\partial^{d_{2}}_{x}\mathcal{G}_{\hat{u}})(\cdot) \mid d_{1}+d_{2} \geq 1; d_{1}, d_{2} = 0,1,2\}$ for transforming to the integral weak forms, $(\Phi(\hat{u}), q_{0})$. Exhaustive all subsets selection solved Eq. (\ref{eq:best_subset}), attaining $\hat{\xi}^{k}$ for every $k \leq N_{q}=8$ (an constant intercept or bias excluded). 

Figure \ref{fig:burgers} (top) shows the BIC scores of the found PDEs, where the support sizes are arranged in increasing order. After the $2$-support PDE, the improvement in BIC becomes stagnant. However, the model selection based on BIC would not choose the $2$-support PDE as the optimal choice. This is because the scores continue decreasing beyond the plateau in the BIC scores, and the $5$-support PDE actually yields the lowest BIC score. We inspect that the log-likelihood dominates the BIC score, when the number of samples in the library $N_{\Omega}$ is large. The evidence reveals the impropriety of using the BIC for identifying the true governing PDE. 

To leverage the proposed UBIC, we quantify the uncertainty $\U^{k}$ from all the best subsets, as plotted in Figure \ref{fig:burgers}. The PDE with $2$ support sizes exhibits the highest stability. These uncertainties are incorporated to penalize the previously obtained BIC scores, preferring the parsimonious PDE with the reliable coefficient estimates. Algorithm \ref{alg:1} suggests the UBIC scores with $\lambda_{\U} = 10^{0} = 1$. As a result, the selected PDE aligns with the true Burgers' PDE form. 

\subsubsection{Korteweg–De Vries (KdV) PDE} We generated the two-soliton $u$ with the initial condition $u(x, 0) = -\sin(\frac{\pi x}{20})$. We employed denoising regularized KSVD with $\rho = 0.01$ on the stack $S_{p}(\Tilde{u})$ created with $25 \times 25$ patches. We set the OMP algorithm with one target sparsity during the training.  

Next, the candidate functions $\phi_{j}(\cdot)$ were chosen from $\{(\mathcal{G}^{d_{1}}_{\hat{u}}\partial^{d_{2}}_{x}\mathcal{G}_{\hat{u}})(\cdot) \mid d_{1}+d_{2} \geq 1; d_{1} = 0,1,2 \text{ and } d_{2} = 0,1,2,3,4\}$ before building their weak forms and $q_{0}$. We exhaustively searched for all the optimal subsets respecting every number of supports, achieving $\forall k \leq N_{q}=14,\, \hat{\xi}^{k}$. 

In Figure \ref{fig:kdv}, we observe that the true equation favored by the UBIC with tuned $\lambda_{\U} = 10^{2.28}$ stands out, in accordance with the minimal uncertainty, than the other potential PDEs. 

\subsubsection{Kuramoto–Sivashinsky (KS) PDE} Following the PDE-FIND paper \cite{rudy2017data}, we experimented with the identical chaotic PDE generated with the initial condition: $u(x, 0) = \cos(\frac{x}{16})(1+\sin(\frac{x}{16}))$. We used the same regularized KSVD settings as detailed in the KdV example. 

Since the set of candidate functions adopted in the KdV example was sufficient to build an overcomplete weak form library for recovering the KS PDE, we underwent the same best-subset regression strategy. 

As shown in Figure \ref{fig:ks}, the lowest BIC and UBIC score with just $\lambda_{\U}=1$ offer the desirable outcome, pointing to the true equation form. 

\begin{table}
    \centering
    \caption{The better is \underline{underlined} (\%CE) or on \textbf{bold} ($R_{\textrm{BIC}}$).}
    \setlength{\tabcolsep}{4pt}
    \begin{tabular}{|c|cc|cc|}
        \hline
        Dataset & \multicolumn{2}{c|}{w/o Denoising} & \multicolumn{2}{c|}{w/ Denoising} \\
         & \%CE & $R_{\textrm{BIC}}$ & \%CE & $R_{\textrm{BIC}}$ \\
        \hline
        Burgers & \underline{$0.7900$} & $-6919$ & $0.9108$ & $\mathbf{-12236}$ \\
        KdV & $17.34$ & $-16614$ & \underline{$9.2987$} & $\mathbf{-22419}$ \\
        KS & $0.4508$ & $-43121$ & \underline{$0.3813$} & $\mathbf{-43533}$ \\
        NS & False Eq. & $-4379570$ & \underline{$11.43$} & $\mathbf{-28710947}$ \\
        \hline
        RD: \makecell[c]{$u_{t}$\\$v_{t}$} & \makecell[c]{$3.1177$\\$3.3251$} & \makecell[c]{$-14790$\\$-15729$} & \makecell[c]{\underline{$2.1639$}\\\underline{$2.2967$}} & \makecell[c]{$\mathbf{-14963}$\\$\mathbf{-15916}$} \\
        \hline
        GS: \makecell[c]{$u_{t}$\\$v_{t}$} & \makecell[c]{\underline{$0.02621$}\\$0.01108$} & \makecell[c]{$-106720$\\$-114432$} & \makecell[c]{$0.05542$\\\underline{$0.01096$}} & \makecell[c]{$\mathbf{-113852}$\\$\mathbf{-120588}$} \\
        \hline
    \end{tabular}
    \label{tab:perf}
\end{table}

\subsubsection{Navier-Stokes (NS) PDE}
We consider the explicit form of the NS equation given in the 3D spatial-temporal grid. As seen in Table \ref{tab:dataset}, $w$ denotes the vorticity. The components of the velocity field are denoted by $u$ ($x$-component) and $v$ ($y$-component), which are both treated as known terms to construct an overcomplete library. We generated the dataset according to the instructions given in the PDE-FIND paper and focused on the bounded spatial domain $(x, y) \in [2, 8.48] \times [0.3, 3.68]$ after the cylinder. We were left with, in total $8,342,750$ data points, for each variable: $w$, $u$, $v$, to which 1\%-sd noise was added after the subsampling. 

To obtain each noise-reduced variable, we applied 2D Savitzky-Golay filters along the temporal axis and then employed the denoising SVD. As experimented in the PDE-FIND, we specifically retained the top singular values, which were $26$, $20$, and $20$ for $w$, $u$, and $v$ (reshaped as metrics with $325 \times 170$ rows and $151$ columns), respectively. The process enhanced the variables' quality, thereby preserving the correct equation form to be captured at the discovered $4$-support PDE. There were $19$ non-weak terms listed from the variables: $\psi_{1} \in \{w, u, v\}$, the spatial derivatives of the vorticity $w$: $\psi_{2} \in \{w_{x}, w_{xx}, w_{y}, w_{yy}\}$ or the interactions $\psi_{1}\psi_{2}$. Every best subset, whose cardinality ranges from $1$ to $10$ is initially approximated by the MIQP (mixed-integer quadratic programming) with the L0-norm based budget constraint \cite{MIO}. Then we carried out an all-subsets exhaustive search over the top candidates, each at least existing in one of the $10$ best subsets. 

Despite the underlying PDE form being dependent on the $4$ supports, it is the most stable one, as illustrated in Figure \ref{fig:ns} (bottom). Undoubtedly, the quantified uncertainty associated with each discovered PDE is a beneficial factor for finding the correct model by the UBIC with $\lambda_{\U}=10^{6.11}$. 

\subsubsection{Denoising Effect} The positive effect of denoising is clearly evident from the major drop in BIC, observed throughout the previously studied examples. We measure the improvement by the maximum reduction in the BIC: $R_{\textrm{BIC}} = \min_{k}\textrm{BIC}(\hat{\xi}^{k}_{\mu})-\max_{k}\textrm{BIC}(\hat{\xi}^{k}_{\mu})$ in Table \ref{tab:perf}. In Figure \ref{fig:ks}, the reduction in BIC scores is not as pronounced as what is demonstrated in the Burgers and KdV examples, implying the challenge of restoring the chaotic pattern of the KS PDE. Meanwhile, in the NS example, the omission of the true PDE when no denoising processes were performed causes the noticeable gap in the BIC values between the two cases, as illustrated in Figure \ref{fig:ns}, confirming the usefulness of the denoising step to the model selection. $R_{\textrm{BIC}}$ or generally the area between trade-offs is a prospective metric for tuning denoising hyperparameters. 

\subsubsection{Robust Adaptive Model Selection}
We fully harness the capability of our proposed method in such extremely noisy scenarios that, without the denoising step, the true governing PDE would be omitted or poorly recovered from the potential best subsets. As visualized in Figure \ref{fig:extreme_noise}, we correctly identify the governing PDEs selected using the adaptive UBIC despite the severe noise interference. Particularly noteworthy is the Burgers example, where neither the BIC nor the uncertainty alone can recover the true equation form. 

\subsubsection{Discovery Accuracy} We evaluate the proximity between the discovered $\hat{\xi}^{k}_{j}$ and ground truth $\xi_{j}$ by the \%CE (percentage coefficient error): $100\times\abs{\hat{\xi}^{k}_{j}-\xi_{j}}/\abs{\xi_{j}}$. The average over every effective coefficient is reported in Table \ref{tab:perf} for each dataset in Table \ref{tab:dataset}. $\hat{\xi}^{k}$ obtained on the denoised data delivers the lower average \%CE than the case without denoising. 

\begin{figure}[t]
\begin{subfigure}{0.48\linewidth}
  \centering
  \includegraphics[width=\textwidth]{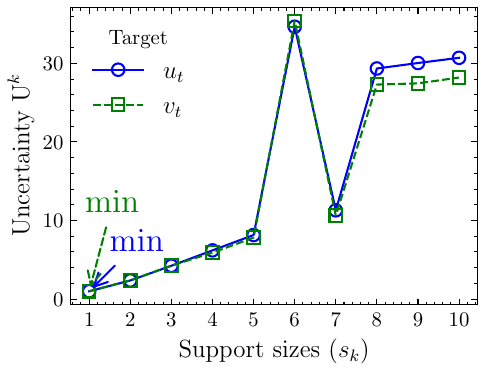}  
  \includegraphics[width=\textwidth]{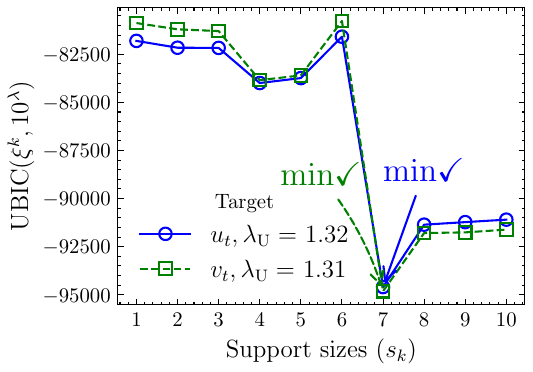}  
  \caption{RD}
  \label{fig:rd2d}
\end{subfigure}
\begin{subfigure}{0.48\linewidth}
  \centering
  \includegraphics[width=\textwidth]{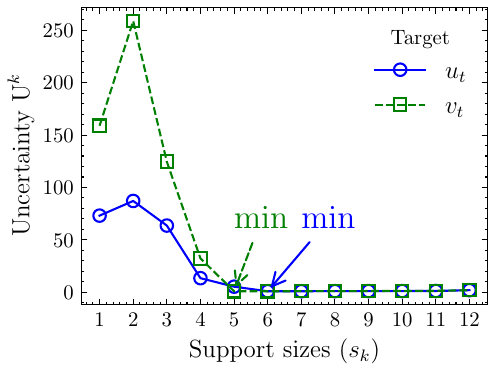}  
  \includegraphics[width=\textwidth]{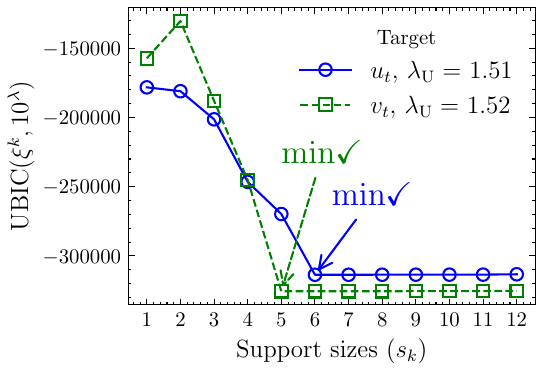}
  \caption{GS}
  \label{fig:rd3d}
\end{subfigure}
\caption{Model selection results for the RD and GS PDEs.}
\label{fig:UBIC_res2}
\end{figure}

\subsubsection{Reaction-Diffusion (RD) PDE}
The PDE governs a system that simulates double spiral waves on a periodic domain, consisting of 7 actual terms. To countermeasure 10\%-sd noise that perturbed a stack of the $u$ and $v$ variables, 2D Savitzky-Golay filters were employed along the temporal axis, the results were then collected to construct the noise-reduced data. 

The candidates were chosen from the variables and their transformations: $\psi_{1} \in \{u, v, u^{3}, v^{3}, u^{2}v, uv^{2}\}$, the spatial derivatives (up to second-order) of either $u$ or $v$: $\psi_{2} \in \{u_{x}, u_{y}, u_{xx}, u_{yy}, u_{xy}, v_{x}, v_{y}, v_{xx}, v_{yy}, v_{xy}\}$, or possible interactions $\psi_{1}\psi_{2}$. To identify the best subset for each cardinality from $1$ to $10$, an initial approximation was obtained using the MIQP with the budget constraint based on the L0-norm. Subsequently, we validated and ensured the optimality of the subsets, whose support sizes were not greater than $10$, respecting the set of unique effective candidates.

Figure \ref{fig:rd2d} shows that the uncertainty alone is not enough for the model selection as it positively correlated with the number of supports, signifying the necessity of the base BIC in Eq. (\ref{eq:UBIC}). The single-support PDE exhibits the least uncertainty. However, an intriguing observation emerges at the $7$-support PDE, where the uncertainty drops relatively to the surrounding PDEs plotted alongside. The stability is effectively exploited by the tuned UBIC with $\lambda \approx 1.31$ to successfully identify the true $7$ candidates. 

\subsubsection{Gray-Scott (GS) PDE}
The GS PDE models the reaction-diffusion system in the 4D spatial-temporal grid. We looped over the stack of the variables along the $t$-temporal then $z$-spatial axes to mitigate the noise effect by 2D Savitzky-Golay filters. Hereafter, similarly to the NS example, each variable was reconstructed via the denoising SVD, retaining the $10$ most significant singular values.

We comprised an overcomplete candidate library with the variables (including an intercept) and their transformations: $\{1, u, v, u^{3}, v^{3}, u^{2}v, uv^{2}\}$, the spatial derivatives of $u$: $\{u_{x}, u_{y}, u_{z}, u_{xx}, u_{yy}, u_{zz}, u_{xy}, u_{xz}, u_{yz}\}$, and the spatial derivatives of $v$: $\{v_{x}, v_{y}, v_{z}, v_{xx}, v_{yy}, v_{zz}, v_{xy}, v_{xz}, v_{yz}\}$. The approximated best subsets were retrieved using the FROLS solver with a maximum support size of $12$. These subsets were guaranteed to be at their optimum within all the effective candidates, each once delivered by the solver.

According to Figure \ref{fig:rd3d}, it is evident that the best subsets, which align with the true complexity of the PDE system with support sizes of $6$ and $5$ for the targets $u_{t}$ and $v_{t}$, exhibit the minimal uncertainties. The tuned UBIC ($\lambda \approx 1.51$) leverages the uncertainty pattern to penalize the BIC values, hence the successful identification of the true PDE system. 

\begin{table}
    \begin{center}
    \caption{PINN-based model selection between the PDEs with (optimal) $s_{k^{*}}$ and (sub-optimal) $s_{k^{*}+1}$ support sizes.}
    \setlength{\tabcolsep}{4pt}
    \begin{tabular}{|c|c|c|c|}
        \hline
        Support sizes & Burgers & KdV & KS \\
        \hline
        \makecell[c]{$s_{k^{*}}$\\$s_{k^{*}+1}$} & \makecell[c]{$\mathbf{-16439}(2)$\footnotemark[1]\\$-2020(3)$} & \makecell[c]{$\mathbf{247070}(2)$\\$280799(3)$} & \makecell[c]{$\mathbf{339036}(3)$\\$501440(4)$}\\
        \hline
    \end{tabular}
    \label{tab:sim_ic}
    \end{center}
    \footnotesize\footnotemark[1]{The support sizes are denoted in parentheses.}
\end{table}

\subsubsection{Simulation-based Model Comparison}
We simulated the PDE selected by the tuned UBIC and another PDE with an additional candidate, using the PINN learning. Comparing the two PDEs, we measured the proximity of their simulated solutions to the denoised version of the observed state variable using the BIC in Eq. (\ref{eq:pinn_bic}). Table \ref{tab:sim_ic} warrants that the $s_{k^{*}}$-support PDE has indeed the sufficient complexity in yielding the lower-BIC simulated state variable than its competitor, the $s_{k^{*}+1}$-support PDE with the extra candidate. 
\section{Conclusion}
We extend the BIC to the parameter-adaptive UBIC, which incorporates the model uncertainty for selecting the governing PDE amidst noisy data. The uncertainty, quantified from the model coefficient posterior, promotes the reliability into the model selection, preventing the rise of overfitted PDEs unaddressed by the BIC. The proposed UBIC demonstrates the successful identification of the true hidden equation while maintaining computational efficiency thanks to the analytical posterior form. Validating consistency in model selection results, we perform a comparison between the PDE selected by the UBIC and its competitor with an extra support to find the better PDE that delivers a lower BIC value calculated between the simulated state and the denoised observed data. Notably, the discovery on denoised data positively improves the BIC trade-off. Towards more complicated examples, we will explore applying the UBIC to the evolutionary-based discovery methods \cite{xu2020dlga} to relax the overcompleteness assumption. 
\bibliography{aaai24}
\clearpage

\begin{figure*}[t]
\begin{subfigure}{.331\textwidth}
  \centering
  \includegraphics[width=\linewidth]{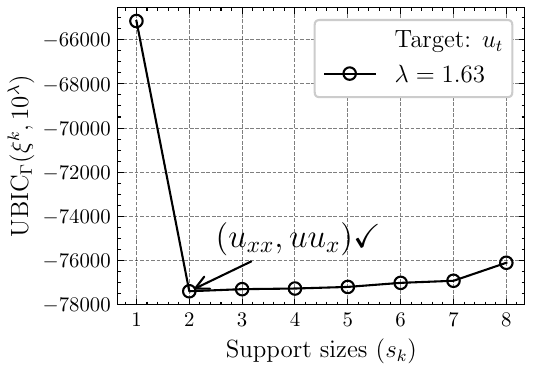}  
  \caption{Burgers}
\end{subfigure}
\begin{subfigure}{.331\textwidth}
  \centering
  \includegraphics[width=\linewidth]{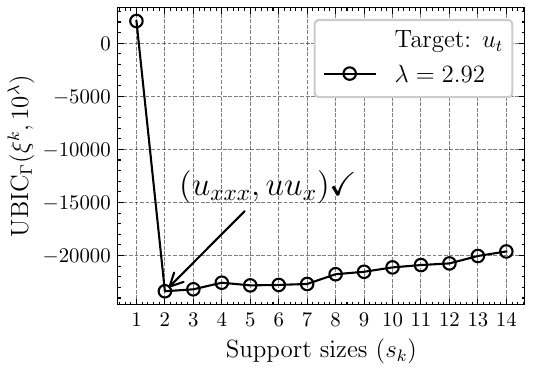}  
  \caption{KdV}
\end{subfigure}
\begin{subfigure}{.331\textwidth}
  \centering
  \includegraphics[width=\linewidth]{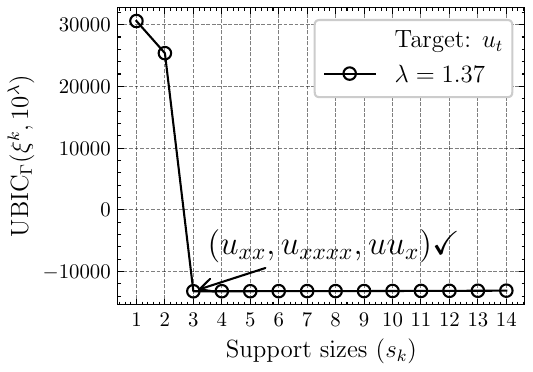}  
  \caption{KS}
\end{subfigure}
\vfill
\begin{subfigure}{.331\textwidth}
  \centering
  \includegraphics[width=\linewidth]{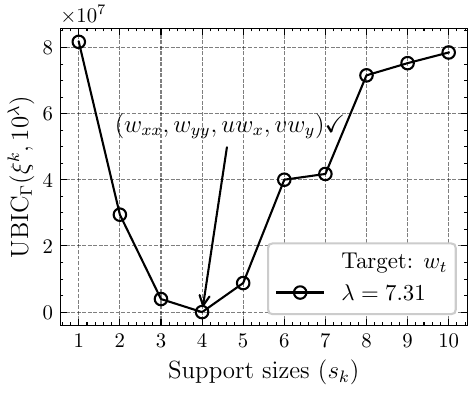}  
  \caption{NS}
\end{subfigure}
\begin{subfigure}{.331\textwidth}
  \centering
  \includegraphics[width=\linewidth]{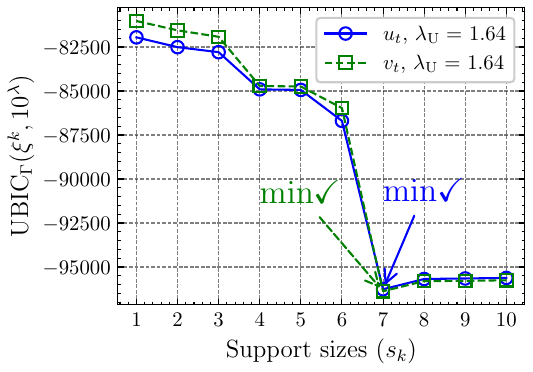}  
  \caption{RD}
\end{subfigure}
\begin{subfigure}{.331\textwidth}
  \centering
  \includegraphics[width=\linewidth]{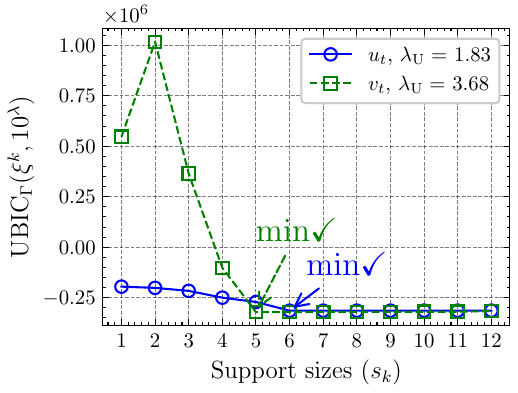}  
  \caption{GS}
\end{subfigure}
\caption{Successful model selection results by G-BIC with $\Gamma = 1$. In each subfigure, the (re)tuned $\lambda$(s) is/are annotated. It should be noted all the $\U^{k}$s are already given in Figure (\ref{fig:UBIC_res1}) and Figure (\ref{fig:UBIC_res2}) in the main text.}
\label{fig:GBIC_res}
\end{figure*}

\begin{figure*}
\begin{subfigure}{.331\textwidth}
  \centering
  \includegraphics[width=\linewidth]{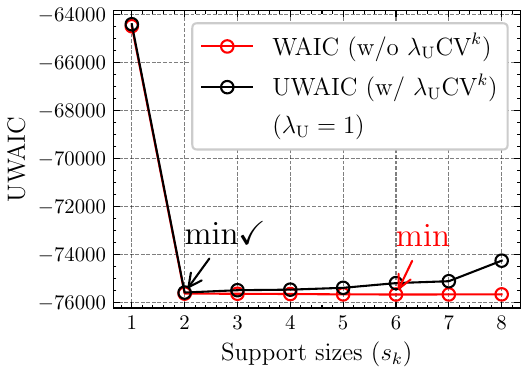}  
  \caption{Burgers}
\end{subfigure}
\begin{subfigure}{.331\textwidth}
  \centering
  \includegraphics[width=\linewidth]{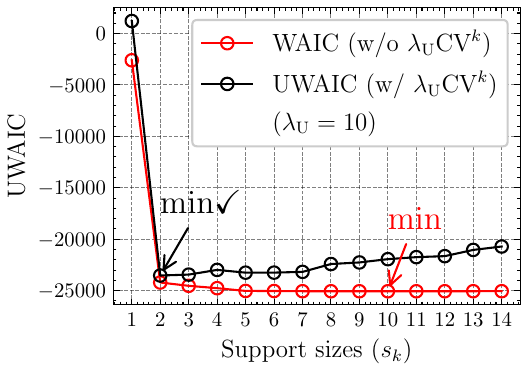}  
  \caption{KdV}
\end{subfigure}
\begin{subfigure}{.331\textwidth}
  \centering
  \includegraphics[width=\linewidth]{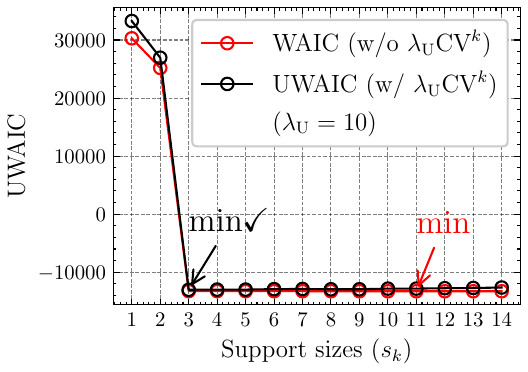}  
  \caption{KS}
\end{subfigure}
\caption{Extension to the uncertainty-penalized WAIC (UWAIC)}
\label{fig:UWAIC_res}
\end{figure*}

\begin{figure*}
\begin{subfigure}{.245\textwidth}
  \centering
  \includegraphics[width=\linewidth]{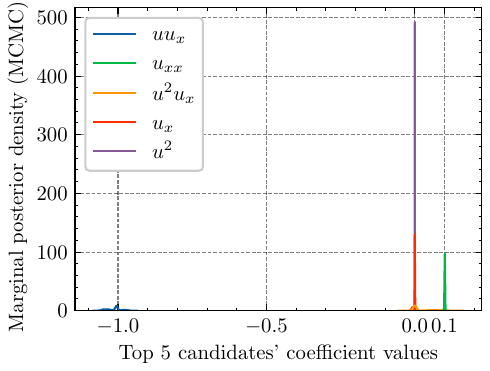}  
  \caption{Marginal posterior}
\end{subfigure}
\begin{subfigure}{.245\textwidth}
  \centering
  \includegraphics[width=\linewidth]{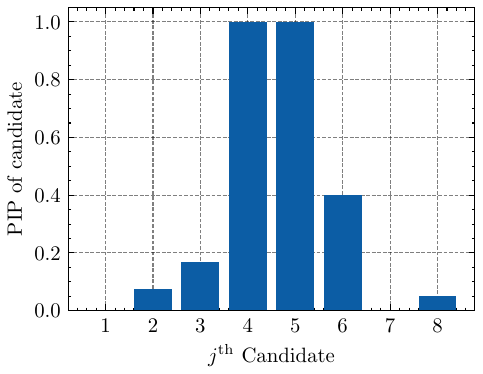}  
  \caption{Candidates' PIP}
\end{subfigure}
\begin{subfigure}{.245\textwidth}
  \centering
  \includegraphics[width=\linewidth]{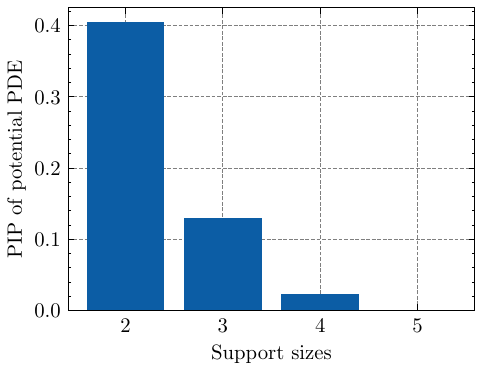}  
  \caption{Best subsets' PIP}
\end{subfigure}
\begin{subfigure}{.245\textwidth}
  \centering
  \includegraphics[width=\linewidth]{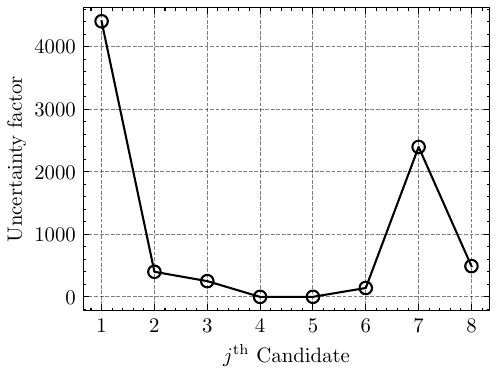}  
  \caption{Uncertainty factor}
\end{subfigure}
\caption{Sparse Bayesian regression with the SS prior ($\beta = 0.3125$) for discovering the Burgers' PDE. The candidates are listed in the following order: $[u, u^{2}, u_{x}, u_{xx}, uu_{x}, u^{2}u_{x}, uu_{xx}, u^{2}u_{xx}]$.}
\label{fig:burgers_ss}
\end{figure*}

\begin{figure*}
\begin{subfigure}{.245\textwidth}
  \centering
  \includegraphics[width=\linewidth]{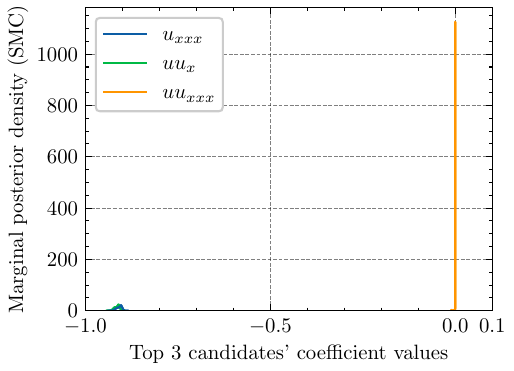}  
  \caption{Marginal posterior}
\end{subfigure}
\begin{subfigure}{.245\textwidth}
  \centering
  \includegraphics[width=\linewidth]{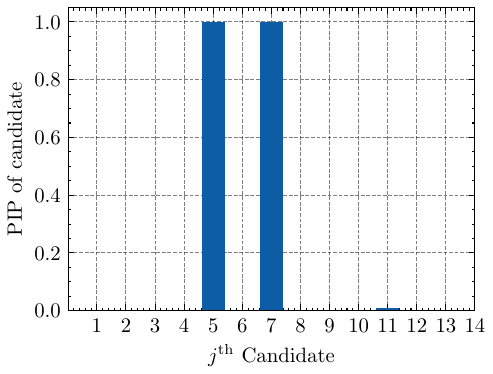}  
  \caption{Candidates' PIP}
\end{subfigure}
\begin{subfigure}{.245\textwidth}
  \centering
  \includegraphics[width=\linewidth]{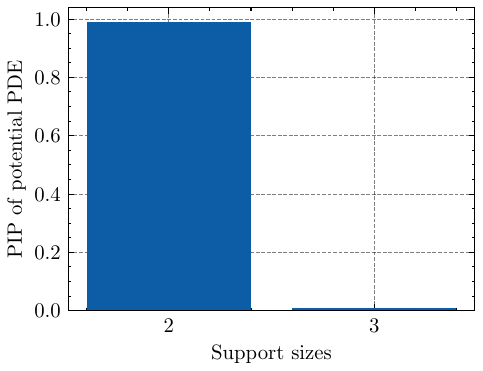}  
  \caption{Best subsets' PIP}
\end{subfigure}
\begin{subfigure}{.245\textwidth}
  \centering
  \includegraphics[width=\linewidth]{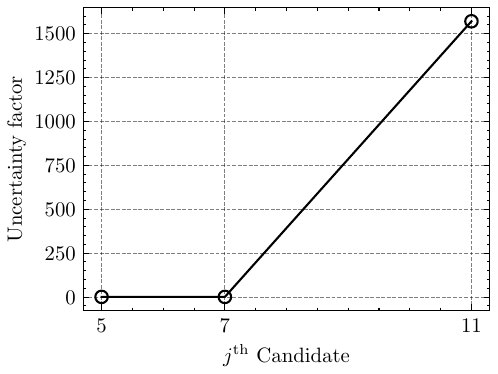}  
  \caption{Uncertainty factor}
\end{subfigure}
\caption{Sparse Bayesian regression with the SS prior ($\beta = 0.01$) for discovering the KdV PDE. The candidates are listed in the following order: $[u, u^{2}, u_{x}, u_{xx}, u_{xxx}, u_{xxxx}, uu_{x}, u^{2}u_{x}, uu_{xx}, u^{2}u_{xx}, uu_{xxx}, u^{2}u_{xxx}, uu_{xxxx}, u^{2}u_{xxxx}]$.}
\label{fig:kdv_ss}
\end{figure*}

\begin{figure*}
\begin{subfigure}{.245\textwidth}
  \centering
  \includegraphics[width=\linewidth]{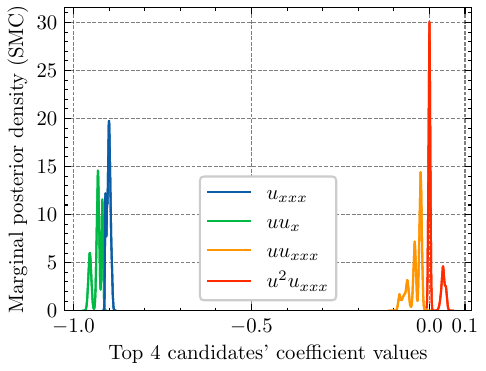}  
  \caption{Marginal posterior}
\end{subfigure}
\begin{subfigure}{.245\textwidth}
  \centering
  \includegraphics[width=\linewidth]{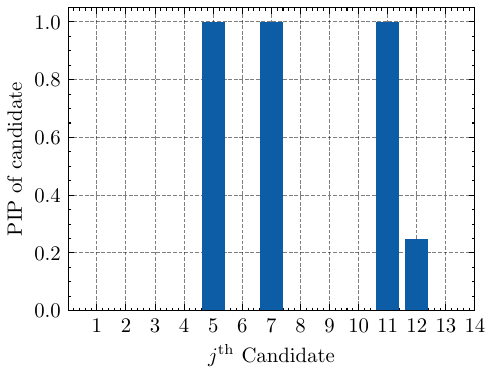}  
  \caption{Candidates' PIP}
\end{subfigure}
\begin{subfigure}{.245\textwidth}
  \centering
  \includegraphics[width=\linewidth]{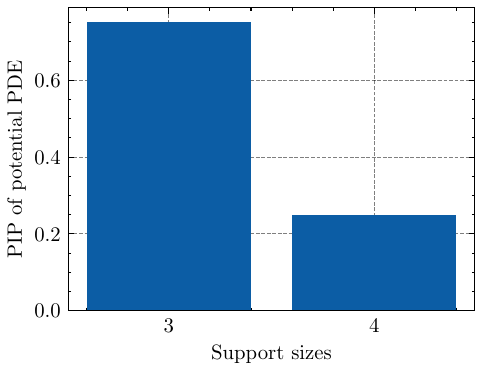}  
  \caption{Best subsets' PIP}
\end{subfigure}
\begin{subfigure}{.245\textwidth}
  \centering
  \includegraphics[width=\linewidth]{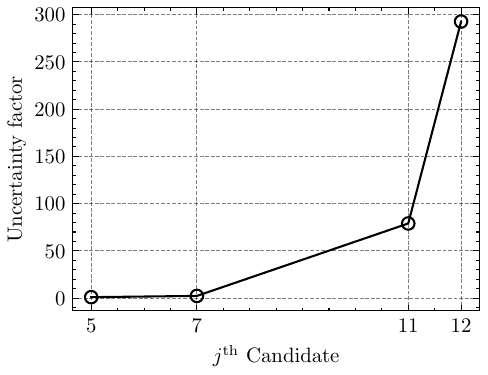}  
  \caption{Uncertainty factor}
\end{subfigure}
\caption{Sparse Bayesian regression with the SS prior ($\beta = 0.1$) for discovering the KdV PDE. Please refer to the candidate order listed in Figure \ref{fig:kdv_ss}.}
\label{fig:kdv_failed_ss}
\end{figure*}

\begin{figure*}
\begin{subfigure}{.48\textwidth}
  \centering
  \includegraphics[width=\linewidth]{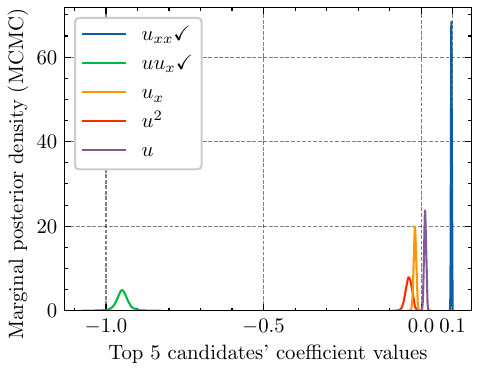}  
  \includegraphics[width=\linewidth]{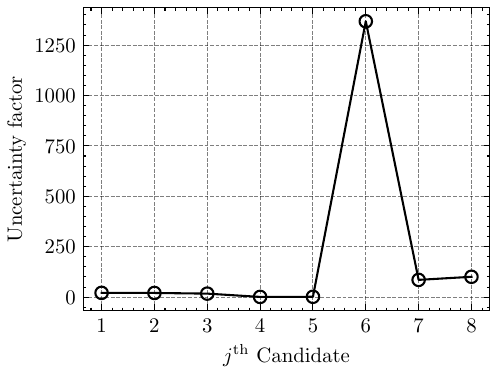}  
  \caption{Burgers}
\end{subfigure}
\begin{subfigure}{.48\textwidth}
  \centering
  \includegraphics[width=\linewidth]{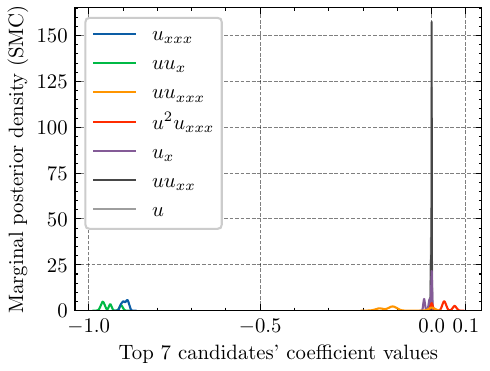}  
  \includegraphics[width=\linewidth]{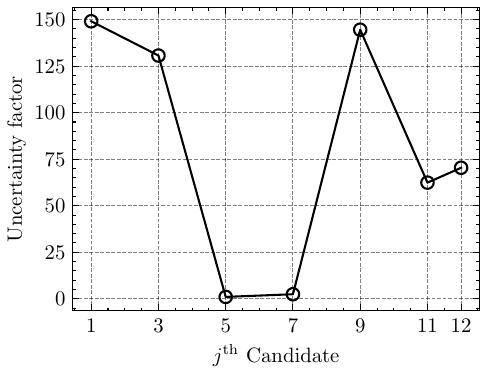}  
  \caption{KdV}
\end{subfigure}
\caption{Sparse Bayesian regression with the RH prior for discovering the Burgers and KdV PDE. The global shrinkage parameter is set equal to $10^{-3}$. Please refer to the candidate order listed in Figures \ref{fig:burgers_ss} and \ref{fig:kdv_ss}.}
\label{fig:rh_res}
\end{figure*}

\begin{figure*}
\begin{subfigure}{.245\textwidth}
  \centering
  \includegraphics[width=\linewidth]{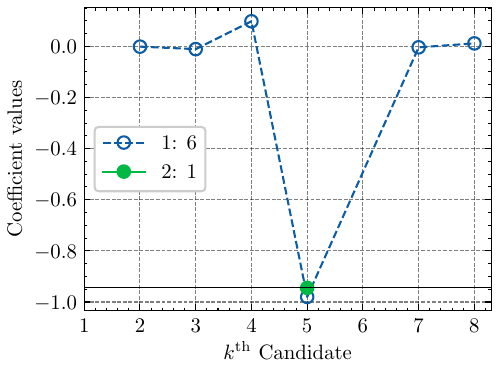}  
  \caption{Threshold $= 10^{-1}$}
\end{subfigure}
\begin{subfigure}{.245\textwidth}
  \centering
  \includegraphics[width=\linewidth]{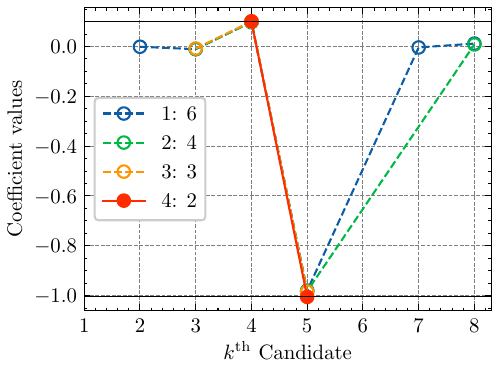}  
  \caption{Threshold $= 10^{-2}$}
\end{subfigure}
\begin{subfigure}{.245\textwidth}
  \centering
  \includegraphics[width=\linewidth]{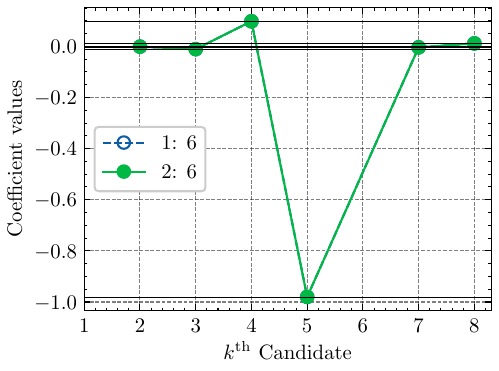}  
  \caption{Threshold $= 10^{-3}$}
\end{subfigure}
\begin{subfigure}{.245\textwidth}
  \centering
  \includegraphics[width=\linewidth]{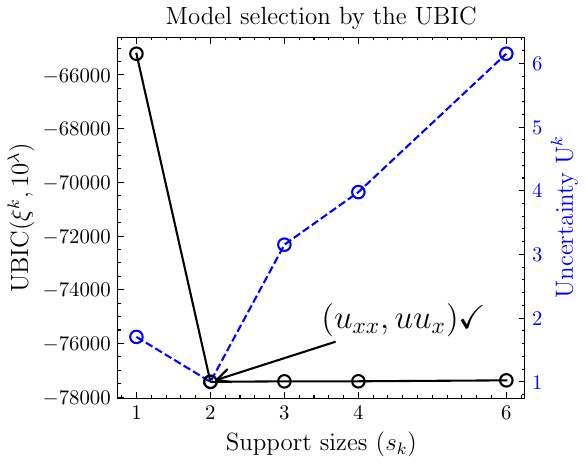}  
  \caption{$\U^{k}$ and UBIC ($\lambda = 0.0$)}
\end{subfigure}
\caption{Threshold sparse Bayesian regression for discovering Burgers' PDE. Please refer to the candidate order listed in Figures \ref{fig:burgers_ss}.}
\label{fig:ardr_res}
\end{figure*}

\section{Supplemental Material}
\subsection{Generalized UBIC (G-UBIC)}
The generalized UBIC (G-UBIC) is defined with a $\Gamma$ scaling parameter as follows:

\begin{equation}
\begin{aligned}
    \textrm{UBIC}_{\Gamma}(\xi^{k}, \lambda_{\U}) &= \textrm{BIC}(\xi^{k}_{\mu}) + \lambda_{\U}\Gamma\U^{k}\\
    &= -2\log L(\xi^{k}_{\mu}) + \log(N_{\Omega})s_{k} + \lambda_{\U}\Gamma\U^{k}, 
\end{aligned}
\end{equation}

\noindent where $\Gamma = \log(N_{\Omega})$ results in Eq. (\ref{eq:UBIC}). $\lambda_{\U}\Gamma$ determines the uncertainty $\U^{k}$ influence on the model selection process. If the influence was given, we would easily find $\lambda_{\U}\Gamma = \lambda^{0}_{\U}\log(N_{\Omega})$, supposing that $\lambda^{0}_{\U}$ were the values presented in the main text. Actually, it is not necessary to keep the influence unvaried, since we only need the appropriate model selection results, which are automatically taken care of by the Algorithm \ref{alg:1} that iteratively checks the likely optimal support sizes (obtained by the G-UBIC) based on the BIC improvement if we transit to the corresponding PDE model. The bound on the maximum $\lambda_{\U}$, which preserves the impact of the log-likelihood on deciding the best PDE, is calibrated according to $\Gamma$. 

\begin{equation}
    \lambda^{\textrm{max}}_{\U} = \max_{k}\frac{2\log\hat{L}(\xi^{k}_{\mu})-\log(N_{\Omega})s_{k}}{\Gamma\U^{k}}
\end{equation}

\noindent The bound is reduced to Eq. (\ref{eq:bound}) with $\Gamma = \log(N_{\Omega})$. As seen in Figure \ref{fig:GBIC_res}, we test the adaptability of Algorithm \ref{alg:1} in tuning $\lambda_{\U} = 10^{\lambda}$ for the basic G-UBIC with $\Gamma = 1$. The correct identification of the true governing form is shown for every canonical PDE dataset we experimented with. 

\subsection{Uncertainty-penalized WAIC (UWAIC)}
Demonstrating the versatility of the core proposals, we conduct a pilot extension of the uncertainty penalization to the WAIC (widely applicable information criterion) \cite{WAIC}, giving $\textrm{UWAIC} = \textrm{T}+\frac{\textrm{FV}}{N_{\Omega}}+\lambda_{\U}\CV^{k}$; where the $\textrm{T}$ and $\textrm{FV}$ are the training loss and functional variance. UWAIC is more computationally expensive than the UBIC or G-UBIC because it involves full Bayesian inference to compute the WAIC before adding the penalizing unnormalized uncertainty $\CV^{k}$. The UWAIC with a $\lambda_{\U}$ specified in Figure \ref{fig:UWAIC_res} is capable of identifying the true PDE terms, similar to the UBIC. These results signify the flexibility of incorporating the proposed uncertainty penalization into other information criteria. 

\subsection{Sparse Bayesian Regression by Sparsifying Priors}
We study the PDE discovery approach based on sparsifying priors as inspired by \cite{UQ-SINDy}. Particularly, we examine sparse Bayesian regression with the spike and slap prior (SS) \cite{ss_prior, madigan1994model, mitchell1988bayesian} and the regularized (Finnish) horseshoe prior (RH) \cite{rh_prior} to solve Eq. (\ref{eq:weak_form}) through Bayesian inference, where we draw samples from the posterior distribution using either Markov chain Monte Carlo (MCMC) or sequential Monte Carlo (SMC) methods implemented in the PyMC package \cite{pymc}. 

For the SS prior each coefficient is given hierarchically as

\begin{equation} \label{eq:ss_prior}
\begin{aligned}
\xi^{\textrm{SS}}_{j} \mid &\mathcal{B}_{j} \sim \mathcal{N}(\hat{\xi}^{k}_{j}, 1)\mathcal{B}_{j},\\
&\mathcal{B}_{j} \sim Bernoulli(\beta);
\end{aligned}
\end{equation}

\noindent where $\beta$ is a probability of success that characterizes the Bernoulli distribution. As a result that $\mathcal{B}_{j} \in \{0, 1\}$, the spike and slap prior enables sparse coefficients. Here, $k$ should be set as the index of the maximum support sizes, depending on available computational resources. 

After completing the Bayesian inference, the quantified uncertainty factor of each coefficient is given by the following calculation over its posterior samples of $\hat{\xi}^{\textrm{SS}}_{j}$: 

\begin{equation}
    \U^{\textrm{SS}}_{j} = \frac{\CV^{\textrm{SS}}_{j}}{\min_{j}{\CV^{\textrm{SS}}_{j}}};\, \CV^{\textrm{SS}}_{j} = \frac{\sqrt{\V[\hat{\xi}^{\textrm{SS}}_{j}]}}{\E[\hat{\xi}^{\textrm{SS}}_{j}]}.
\end{equation}

\noindent Under our core assumption, $\U^{\textrm{SS}}_{j}$ would be comparatively low for the $j^{\textrm{th}}$ effective candidate that corresponds to one of the true terms. For a candidate to be considered effective, its posterior inclusion probability (PIP) must be greater than zero. By counting the number of times a unique subset occurs in the posterior samples, we can also estimate the posterior inclusion probability (PIP) for each (best) subset. The uncertainty factor is computed likewise for the other sparsity-inducing prior (i.e., $\U^{\textrm{RH}}_{j}$ for the RH prior). 

In Figures \ref{fig:burgers_ss} and \ref{fig:kdv_ss}, the desirable outcomes are expected by inspecting that the best subsets with the correct support sizes (2 in these cases) present the highest PIPs. The marginal posterior of the negligible candidates primarily distributes a spike around the origin, whereas wider or slapped distributions are observed for the true nonzero candidates. However, achieving the results comes with the cost of setting the appropriate probability of success of the Bernoulli distribution in Eq. (\ref{eq:ss_prior}), i.e., $\beta = 0.3125$ and $0.01$ respectively for the Burgers and KdV examples. If a bigger $\beta = 0.1$ is asserted for the KdV example, the best subset with the highest PIP is comprised of $3$ candidates instead, which does not convey the true sparsity as shown in Figure \ref{fig:kdv_failed_ss}, hence the troublesome sensitivity caused by $\beta$. On the contrary, the uncertain pattern remains insensitive to the change in $\beta$ from $0.01$ to $0.1$. Also, the $u_{xxx}$ and $uu_{x}$ ($5^{\textrm{th}}$ and $7^{\textrm{th}}$) candidates yield the least uncertainty factor values. The uncertainty factors of $u_{xx}$ and $uu_{x}$ ($4^{\textrm{th}}$ and $5^{\textrm{th}}$ candidates) are found to be minimal for the Burgers' PDE case, illustrated in Figure \ref{fig:burgers_ss}. 

By the RH prior design \cite{UQ-SINDy}, the coefficients sampled from the posterior distribution are not strictly sparse, exhibiting values close to, but not precisely, zero(s). Thus, a definition of pseudo-probabilities may be introduced. In this study, we are inclined to take the uncertainty factor as our preferred alternative. Figure \ref{fig:rh_res} reveals that the true nonzero candidates have the lowest uncertainty factor, akin to the cases where the spike and slap priors are utilized. We set the global shrinkage parameter of the RH prior equal to $10^{-3}$ for both examples. 

\subsection{Threshold Sparse Bayesian Regression}
Another approach for achieving the sparse identification of the governing PDE with quantified uncertainty (error bars) is through iterative thresholding until no further changes in sparsity are detected. This approach is known as threshold sparse Bayesian regression \cite{zhang2018robust}. We adhere closely to their iterative thresholding algorithm. In our implementation, we leveraged fast automatic relevance determination that uses sparse Bayesian learning \cite{tipping2003fast, tipping2001analysis} to estimate mean regression coefficients from the posterior distribution. 

We showcase the experimental results with $3$ different pre-specified threshold values, as depicted in Figure \ref{fig:ardr_res}. We find the sensitivity with respect to the threshold: overly high or low threshold values yield underfitted and overfitted PDE models, respectively. This issue is simply addressed by aggregating the best subsets from the three cases (Threshold $= 10^{-1}, 10^{-2}, 10^{-3}$) and selecting the one that minimizes the UBIC, as seen on the rightmost subfigure. 

\subsection{Improving PDE Discovery Accuracy}
After the model selection is finalized, the coefficient values may be refitted using different candidate representations such as the convolutional weak formulation (CWF). Alternatively, we suggest that the quality of the estimated coefficients can be simply improved by further denoising on $\hat{u}$ over a subdomain $\Omega_{i}$ in Eq. (\ref{eq:weak_form}). Particularly, we specialize $\mathcal{G}_{u^{+}} = \mathcal{S}^{\alpha}_{\Omega_{i}}\circ\mathcal{G}_{\hat{u}}$ instead of using $\mathcal{G}_{\hat{u}}$; where $\mathcal{S}^{\alpha}_{\Omega_{i}}$ is the composite denoising function with $\alpha$ parameterization. Practically, $\mathcal{S}^{\alpha}_{\Omega_{i}}$ is implemented as the Savitzky-Golay filtering with the window size $\alpha$ and applied to the bounded $\hat{u}$ over each $\Omega_{i}$, before computing $\forall j,\, q^{i}_{j}$. Note that the order of the polynomial used for the filtering is $2$. The resulting modification is called the denoised WF (weak formulation). 

The comparison of how close the estimated coefficients are to the true values in terms of the average percentage coefficient error (\%CE) is listed in Table \ref{tab:acc} for the different library representations. With the simple adjustment, the denoised WF effectively outperforms the counterpart CWF method on average over the $3$ PDE datasets. Basically, $\mathcal{S}^{\alpha}_{\Omega_{i}}$ could have been implemented using different denoising filters that may enhance the discovery accuracy even more. Therefore, our work demonstrates not only the effectiveness of the PDE discovery approach using the proposed UBIC but the accurately estimated coefficients. 




\begin{table}
    \centering
    \caption{Comparison based on \%CE between the denoised WF and CWF methods for the library preparation. The lower \%error is on \textbf{bold}.}
    \begin{tabular}{|c|c|c|c|}
        \hline
        Dataset & Denoised WF & CWF on $\hat{u}$ & \makecell[c]{CWF on $\Tilde{u}$\\(undenoised)}\\
        \hline
        Burgers & $\mathbf{0.0237}$ & $1.7287$ & $1.2987$\\
        KdV & $0.36011$ & $1.9493$ & $\mathbf{0.3215}$\\
        KS & $\mathbf{0.2738}$ & $0.3544$ & $0.9935$\\
        \hline
    \end{tabular}
    \label{tab:acc}
\end{table}

\begin{figure*}[t]
\begin{subfigure}{.331\textwidth}
  \centering
  \includegraphics[width=\linewidth]{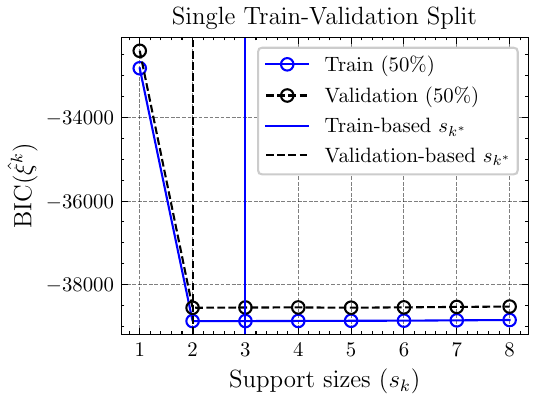}  
\end{subfigure}
\begin{subfigure}{.331\textwidth}
  \centering
  \includegraphics[width=\linewidth]{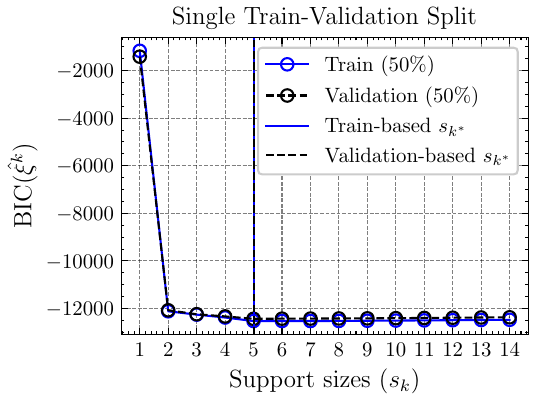}  
\end{subfigure}
\begin{subfigure}{.331\textwidth}
  \centering
  \includegraphics[width=\linewidth]{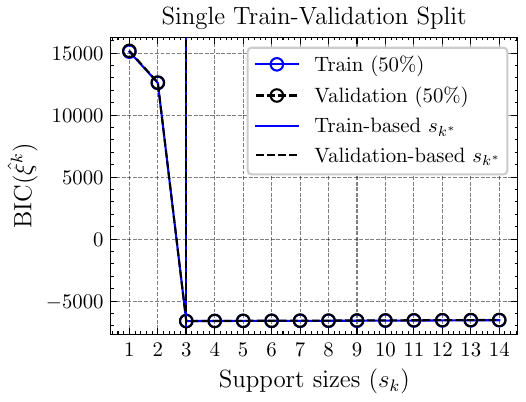}  
\end{subfigure}
\vfill
\begin{subfigure}{.331\textwidth}
  \centering
  \includegraphics[width=\linewidth]{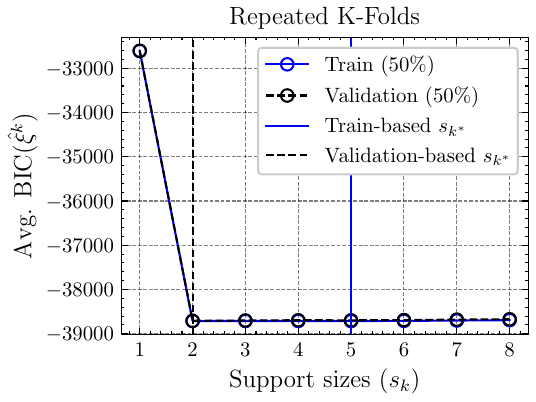}  
\end{subfigure}
\begin{subfigure}{.331\textwidth}
  \centering
  \includegraphics[width=\linewidth]{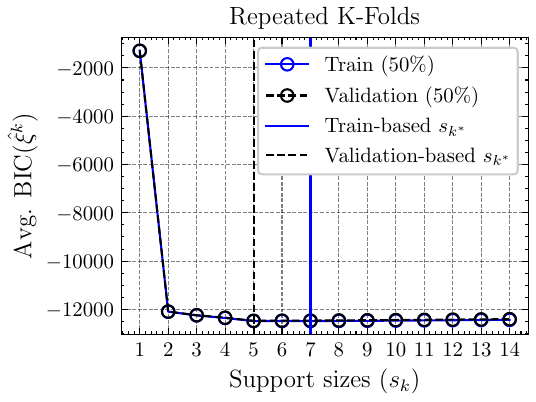}  
\end{subfigure}
\begin{subfigure}{.331\textwidth}
  \centering
  \includegraphics[width=\linewidth]{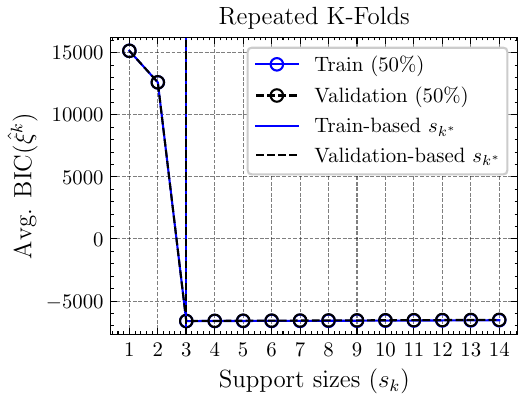}  
\end{subfigure}
\vfill
\begin{subfigure}{.331\textwidth}
  \centering
  \includegraphics[width=\linewidth]{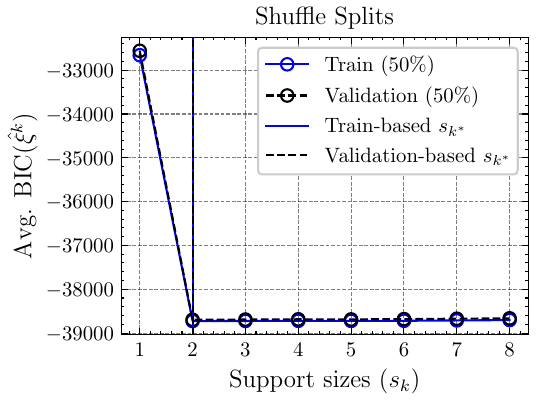}  
  \caption{Burgers}
\end{subfigure}
\begin{subfigure}{.331\textwidth}
  \centering
  \includegraphics[width=\linewidth]{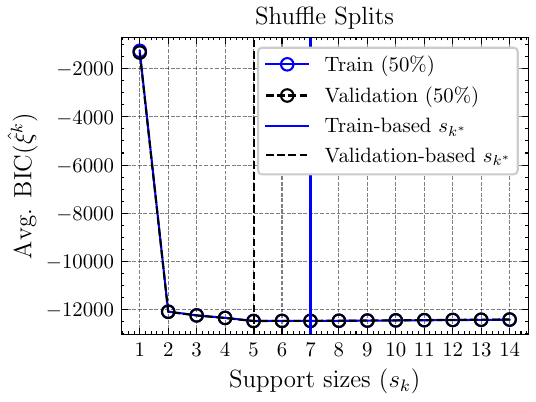}  
  \caption{KdV}
\end{subfigure}
\begin{subfigure}{.331\textwidth}
  \centering
  \includegraphics[width=\linewidth]{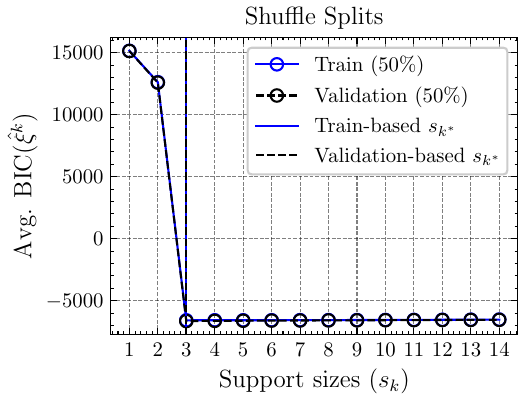}  
  \caption{KS}
\end{subfigure}
\caption{Cross-validation model selection (finding the optimal support sizes) for Burgers, KdV, and KS examples.}
\label{fig:crossval_res}
\end{figure*}

\subsection{Cross-Validation Pitfalls}
One might intuitively presume that the cross-validation strategy would suffice for selecting the optimal support sizes from Eq. (\ref{eq:best_subset}). To validate this assumption, we conducted experiments using various data splitters available in the Scikit-learn package \cite{scikit-learn}. Specifically, we tested the following splitter classes, which are represented respectively in the middle and bottom rows of Figure \ref{fig:crossval_res}. 

\begin{itemize}
    \item \texttt{RepeatedKFold(n\_splits=2, n\_repeats=5)}
    \item \texttt{ShuffleSplit(n\_splits=2, n\_repeats=10)}
\end{itemize}

\noindent The former splitter was configured with 2-folds and repeated 5 times with different randomization in each repetition. Contrarily, the latter splitter was configured with 10 re-shuffling and splitting iterations but did not guarantee that all folds are different. We employed the single train-validation split procedure, as depicted in the first row of Figure \ref{fig:crossval_res}, wherein the first half of the samples were allocated for training and the second half for validation. 

In the Burgers and KS cases, the utilization of validation sets safeguards against the wrong selection of the overfitted PDEs. Nevertheless, none of the splitters were found effective for the KdV example, where there are the noticeable BIC drops observed during the transition from the $2$-support PDE to the $5$-support PDE. We label this scenario as the ``cross-validation pitfalls'', which could have led us to the misconception that the $5$ support sizes are optimal. 

\begin{figure*}[t]
\begin{subfigure}{\textwidth}
  \centering
  \includegraphics[width=\linewidth]{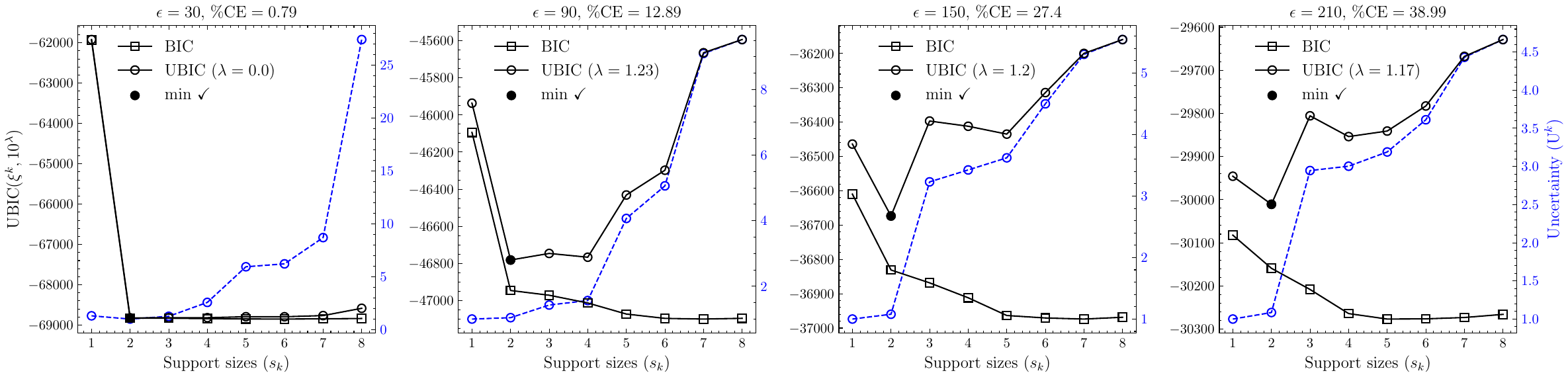}  
  \caption{Burgers}
\end{subfigure}
\begin{subfigure}{\textwidth}
  \centering
  \includegraphics[width=\linewidth]{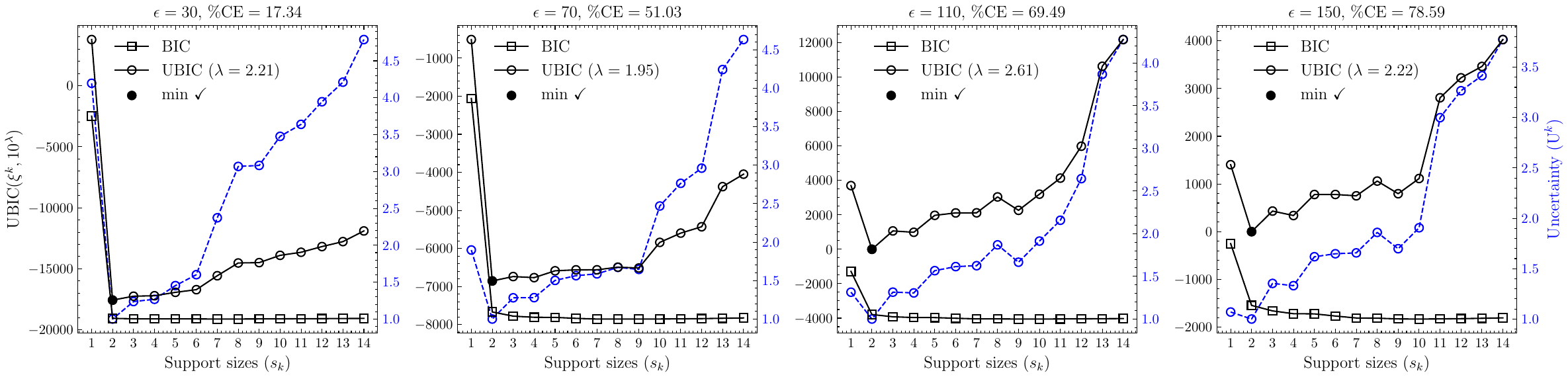}  
  \caption{KdV}
\end{subfigure}
\begin{subfigure}{\textwidth}
  \centering
  \includegraphics[width=\linewidth]{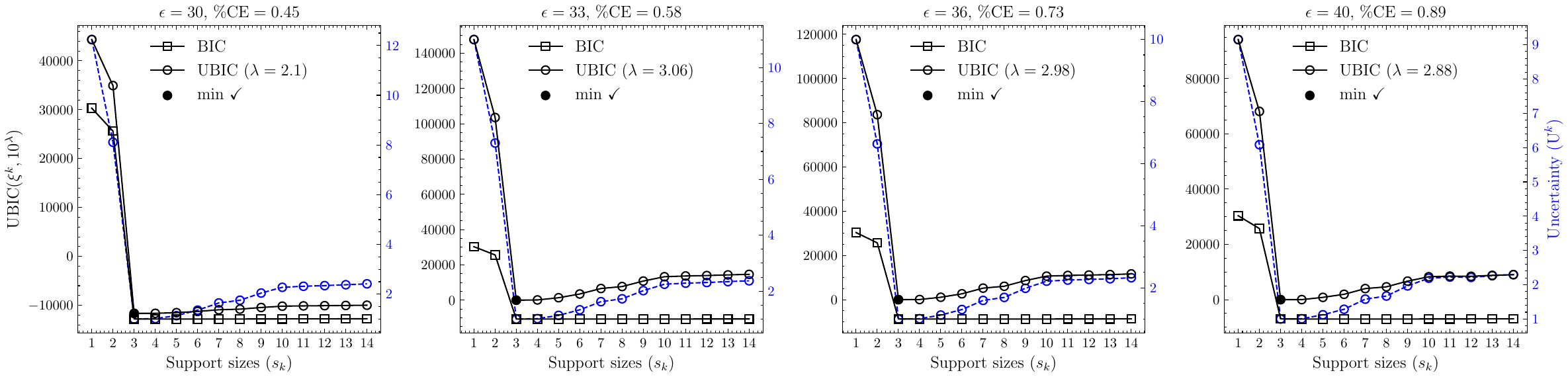}  
  \caption{KS}
\end{subfigure}
\caption{Robust model selection by the UBIC without preceding denoising under the highly noisy scenarios. For the Burgers and KS cases, we assign $\tau_{0} = P_{75}(S)$. For the KdV cases in this Figure, we assign $\tau_{0} = P_{85}(S)$.}
\label{fig:noise_robustness}
\end{figure*}

\subsection{Robust Adaptive Model Selection}
We evaluate the feasibility of our proposed methods in highly noisy scenarios. We ran through the Burgers, KdV, and KS examples with higher levels of noise but, this time, we ablated the denoising step to first justify solely the usability of the UBIC. Surprisingly, we realized that using just the weak formulation and selecting based on the adaptive UBIC, we can truly identify the governing PDEs even under the very high noise levels, as illustrated in Figure \ref{fig:noise_robustness}. The outcomes emphasize the advantage of incorporating penalizing uncertainty information for the model selection, especially in the scenarios contaminated with substantial noise. 

\begin{table}
    \begin{center}
    \caption{Simulation-based model comparison between the two likely PDEs, whose support sizes are $s_{k^{*}}$ and $s_{k^{*}+1}$.}
    \setlength{\tabcolsep}{3.5pt}
    \begin{tabular}{|c|c|c|c|}
        \hline
        Dataset & PINN\footnotemark[1] & Dedalus & Chebfun \\
        \hline
        Burgers \makecell[r]{$s_{k^{*}}=2$\\$s_{k^{*}+1}=3$} & \makecell[c]{$\mathbf{-16439}$\\$-2020$} & \makecell[c]{$\mathbf{-134490}$\\$-134150$} & \makecell[c]{$\mathbf{-134490}$\\$-134160$} \\
        \hline
        KdV \makecell[r]{$s_{k^{*}}=2$\\$s_{k^{*}+1}=3$} & \makecell[c]{$\mathbf{247070}$\\$280799$} & \makecell[c]{$\mathbf{-731845}$\footnotemark[2]\\$-722890$} & \makecell[c]{$\mathbf{-708492}$\footnotemark[2]\\Divergence\footnotemark[3]} \\
        \hline
        KS \makecell[r]{$s_{k^{*}}=3$\\$s_{k^{*}+1}=4$} & \makecell[c]{$\mathbf{339036}$\\$501440$} & \makecell[c]{$\mathbf{783459}$\\$811910$} & \makecell[c]{$\mathbf{819638}$\\$819650$} \\
        \hline
    \end{tabular}
    \label{tab:sim_ic2}
    \end{center}
    \footnotesize\footnotemark[1]{The simulated solution by PINN is evaluated on the validation set $\mathcal{D}_{\textrm{Val}}$ detailed in Table \ref{tab:sim_domain}, unlike the full-domain simulated solutions by Dedalus or Chebfun, which get evaluated on the entire domain.}
    \footnotesize\footnotemark[2]{The $2$-support PDE is added with an intercept and refitted using CWS before the simulation.}
    \footnotesize\footnotemark[3]{The solution by the spin function explodes (diverges) with a small time-step of $10^{-5}$.}
\end{table}
\begin{table*}
    \begin{center}
    \caption{Nonoverlapping train and validation domains particularly bounded for performing the PINN-based model selection.}
    \setlength{\tabcolsep}{4pt}
    \begin{tabular}{|c|cc|cc|}
        \hline
        Dataset & \multicolumn{2}{c|}{$\mathcal{D}_{\textrm{Train}}$} & \multicolumn{2}{c|}{$\mathcal{D}_{\textrm{Val}}$} \\
         & $x$ & $t$ & $x$ & $t$ \\
        \hline
        Burgers & $[-8, 0]$ & $[0, 5]$ & $[0.0625, 7.9375]$ & $[5.1, 10]$ \\
        KdV & $[-20, 19.84]$\footnotemark[1] & $[20.08, 40]$ & $[-19.92, 19.92]$\footnotemark[2] & $[0, 20]$ \\
        KS & $[0.0982, 50.36]$ & $[0, 50]$ & $[50.46, 100.53]$ & $[50.4, 100]$ \\
        \hline
    \end{tabular}
    \label{tab:sim_domain}
    \end{center}
    \centering
    \footnotesize\footnotemark[1]{From even indices of the discretized $x$.} \footnotesize\footnotemark[2]{From odd indices of the discretized $x$.}
\end{table*}

\subsection{Simulation-based Model Selection: Extended Results}
The evidence presented in Table \ref{tab:sim_ic} indicates that the $s_{k^{}}$-support PDE has sufficient complexity to generate the lower-BIC simulated state variable compared to the $s_{k^{}+1}$-support PDE, as demonstrated via the PINN learning. To strengthen the results, we also solve the PDEs on their entire spatio-temporal domain using the Chebfun and Dedalus software, unlike the PINN approach that necessitates a train-validation data split. The PySR package \cite{cranmer2023interpretable} is used to recover the analytical equation of the initial condition. The results are given in Table \ref{tab:sim_ic2}, revealing the same conclusion. 

\begin{figure*}[t]
\begin{subfigure}{.331\textwidth}
  \centering
  \includegraphics[width=\linewidth]{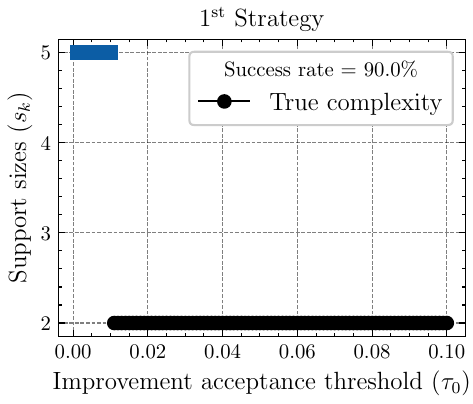}  
\end{subfigure}
\begin{subfigure}{.331\textwidth}
  \centering
  \includegraphics[width=\linewidth]{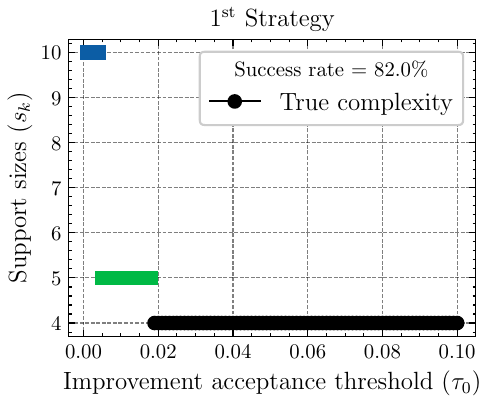}  
\end{subfigure}
\begin{subfigure}{.331\textwidth}
  \centering
  \includegraphics[width=\linewidth]{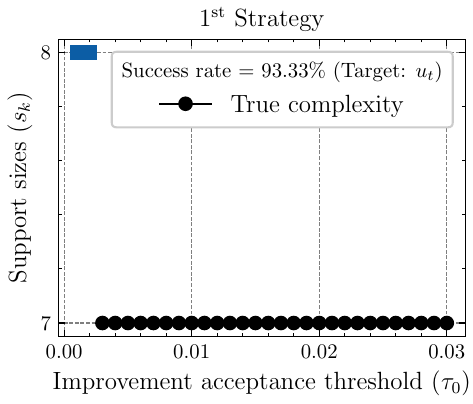}  
\end{subfigure}
\vfill
\begin{subfigure}{.331\textwidth}
  \centering
  \includegraphics[width=\linewidth]{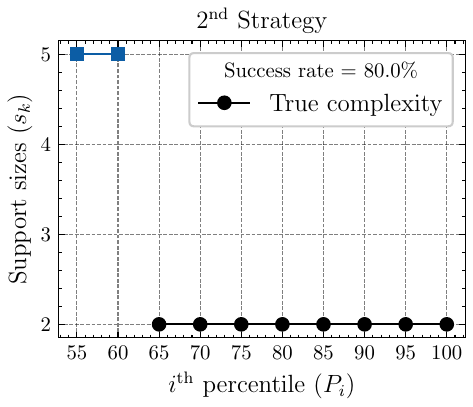}  
  \caption{KdV}
\end{subfigure}
\begin{subfigure}{.331\textwidth}
  \centering
  \includegraphics[width=\linewidth]{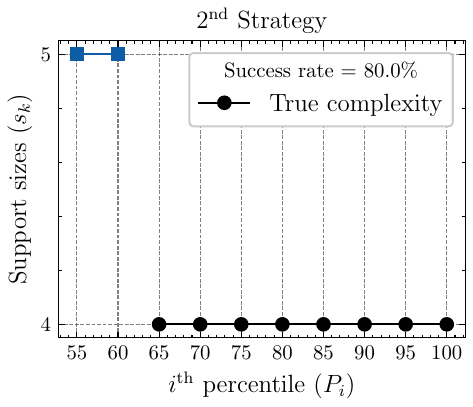}  
  \caption{NS}
\end{subfigure}
\begin{subfigure}{.331\textwidth}
  \centering
  \includegraphics[width=\linewidth]{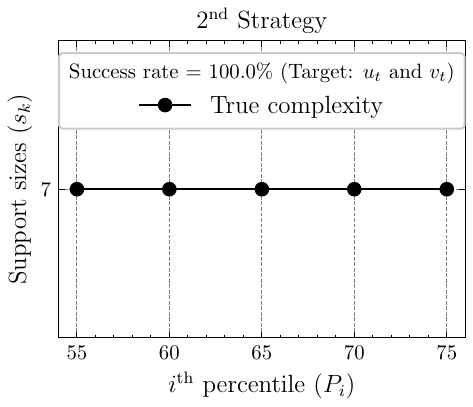}  
  \caption{RD}
\end{subfigure}
\caption{Sensitivity analysis on the KdV, NS, and RD examples, where one (or both) of the success rates of the strategies falls below $100$\%.}
\label{fig:sens_res}
\end{figure*}

\subsection{Sensitivity Analysis}
In this analysis, we assess the sensitivity of the proposed Algorithm \ref{alg:1} by its success rate in identifying the true equation under different values of $\tau_{0}$. Recall that we mention the two distinct strategies for assigning $\tau_{0}$ values that result in the selection of PDEs whose support sizes are greater than one: (i) using raw numerical values and (ii) adopting percentiles of successive improvement factors $S$ (refer to the main text), considering only BIC-decreasing models. The strategies, on which we mainly focus, are given in what follows. 

\begin{itemize}
    \item (i): $\tau_{0} = 10^{-3}(1+h);$ where $h \in \{0, 1, 2, \dots 99\}$.
    \item (ii): $\tau_{0} = P_{i}(S)$; where $i \in \{55, 60, 65, \dots 100\}$.\\$P_{i}$ calculates the $i^{\textrm{th}}$ percentile of the set $S$.
\end{itemize}

\noindent The success rate of each strategy is quantified as the number of times we successfully identify the true equation form over the number of all $\tau_{0}$ values that lead to the selection of PDEs whose support sizes ($s_{k}$) is more than one (to avoid overly high values of $\tau_{0}$). 

We achieve the perfect $100$\% success rate for some of the examples listed in Table \ref{tab:dataset}. For the examples, in which the success rates are less than $100$\% using one of the strategies, we create Figure, showing the suggested support sizes (by Algorithm \ref{alg:1}) against the $\tau_{0}$ values within the specified range. In spite of the possible inappropriate uses of excessively low $\tau_{0}$ that causes the selection of overfitted PDEs, the success rates exceeding $80$\% are deemed acceptable, implying the manageably low sensitivity of the Algorithm \ref{alg:1}. 

\begin{figure*}[t]
\begin{subfigure}{.245\textwidth}
  \centering
  \includegraphics[width=\linewidth]{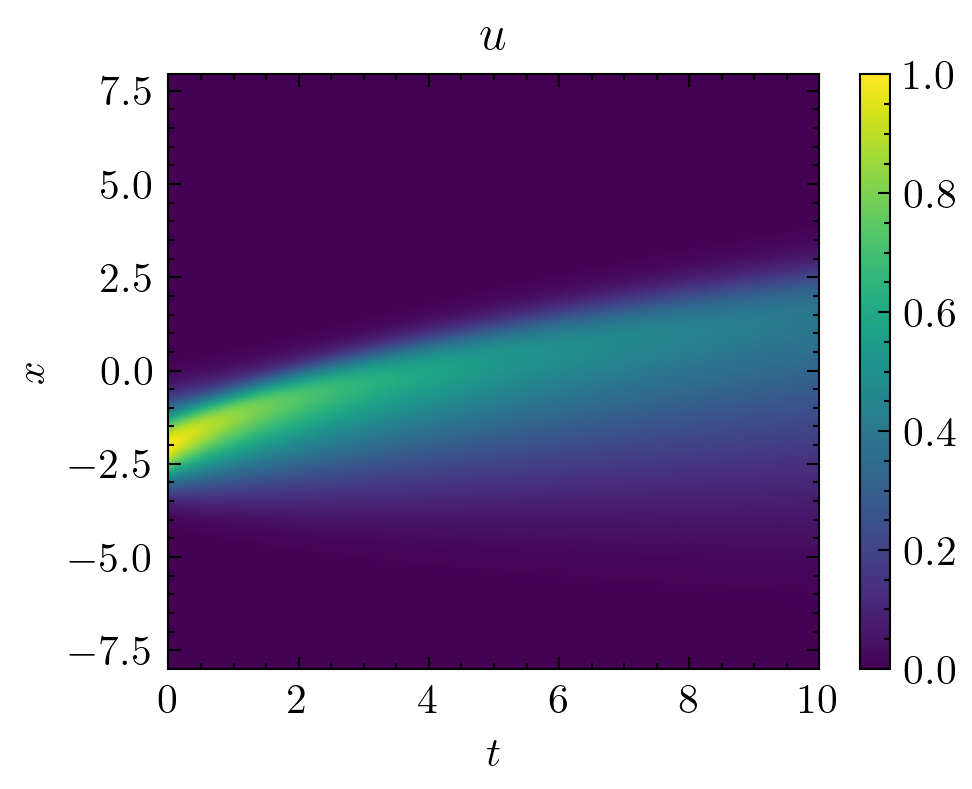}  
  \caption{Burgers}
\end{subfigure}
\begin{subfigure}{.245\textwidth}
  \centering
  \includegraphics[width=\linewidth]{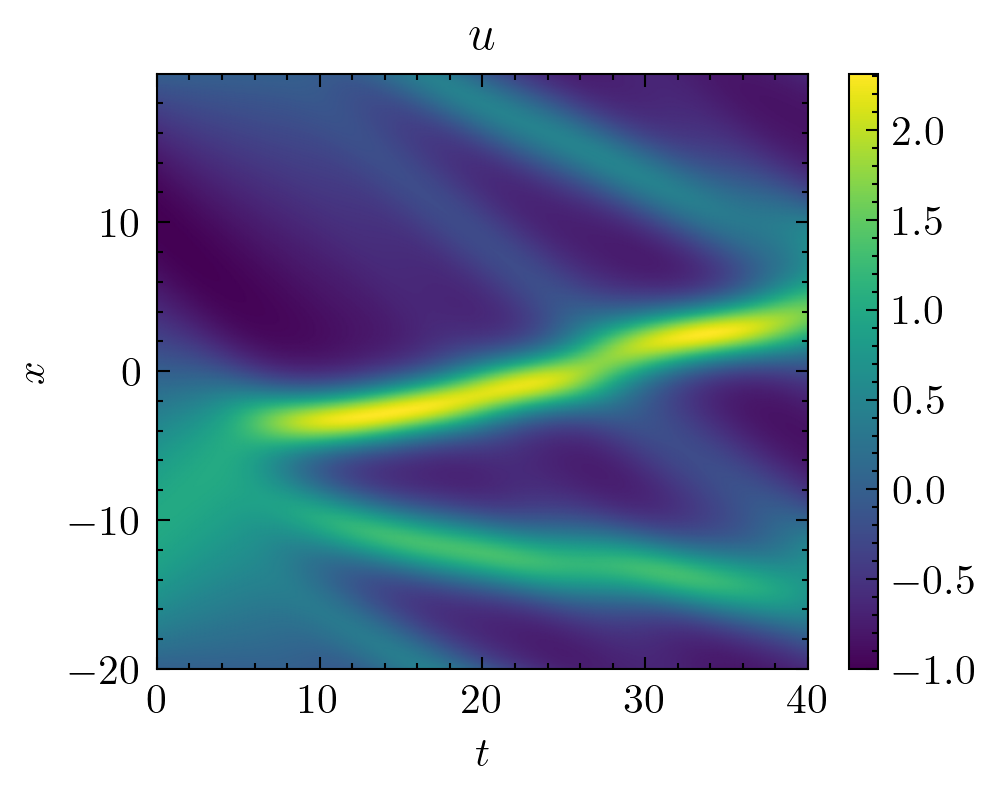}  
  \caption{KdV}
\end{subfigure}
\begin{subfigure}{.245\textwidth}
  \centering
  \includegraphics[width=\linewidth]{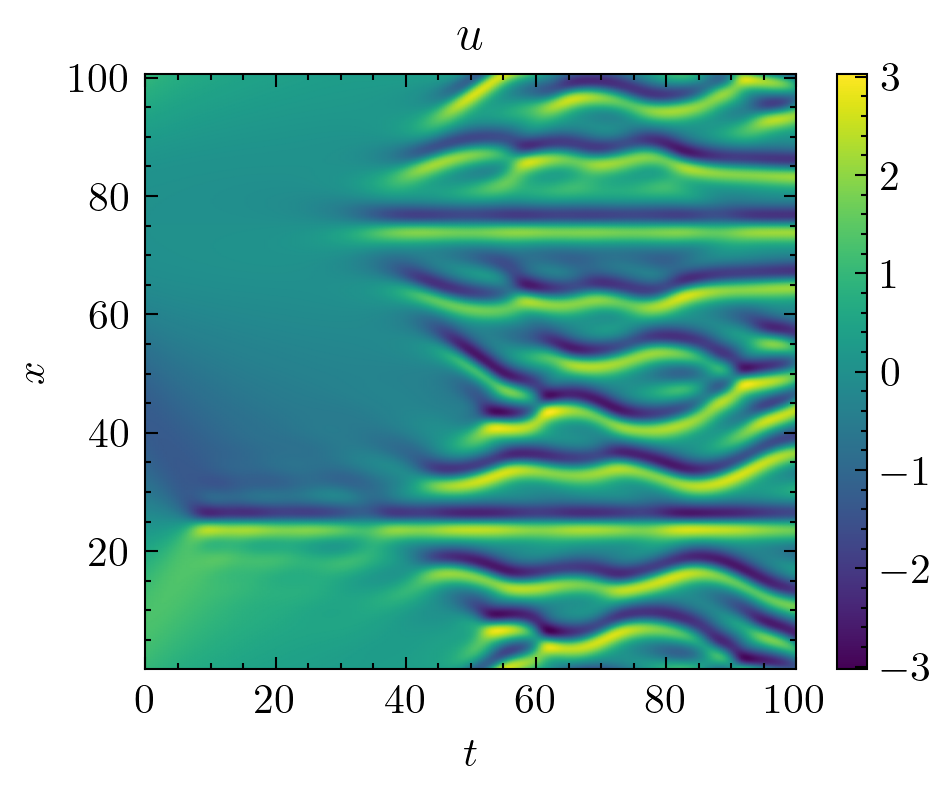}  
  \caption{KS}
\end{subfigure}
\begin{subfigure}{.245\textwidth}
  \centering
  \includegraphics[width=\linewidth]{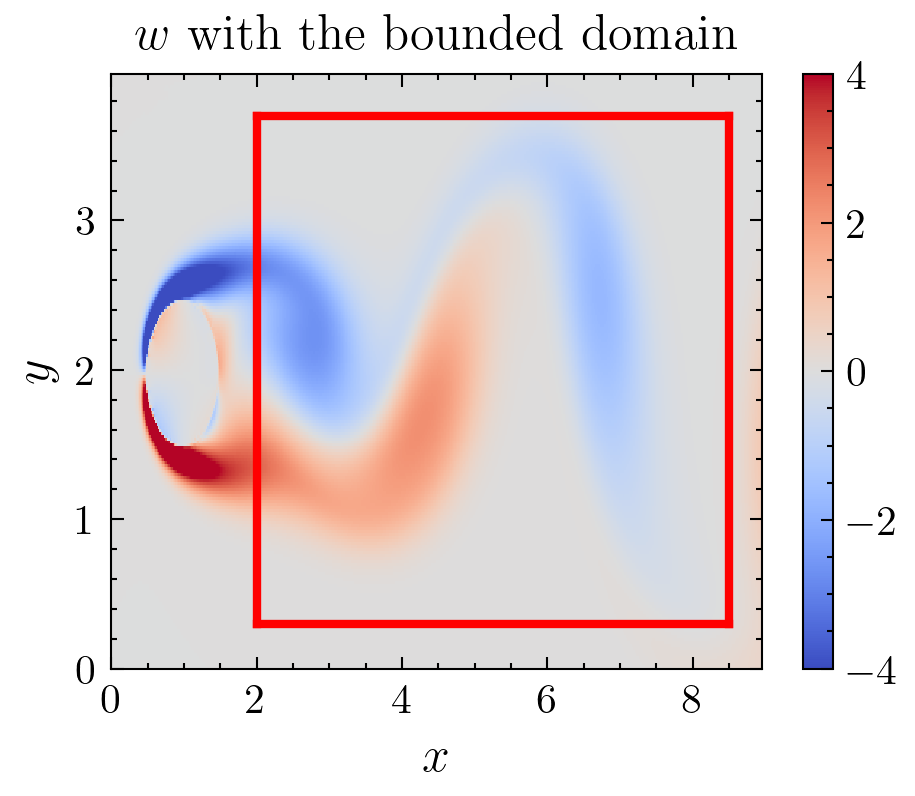}  
  \caption{NS}
\end{subfigure}
\vfill
\begin{subfigure}{.49\textwidth}
  \centering
  \includegraphics[width=\linewidth]{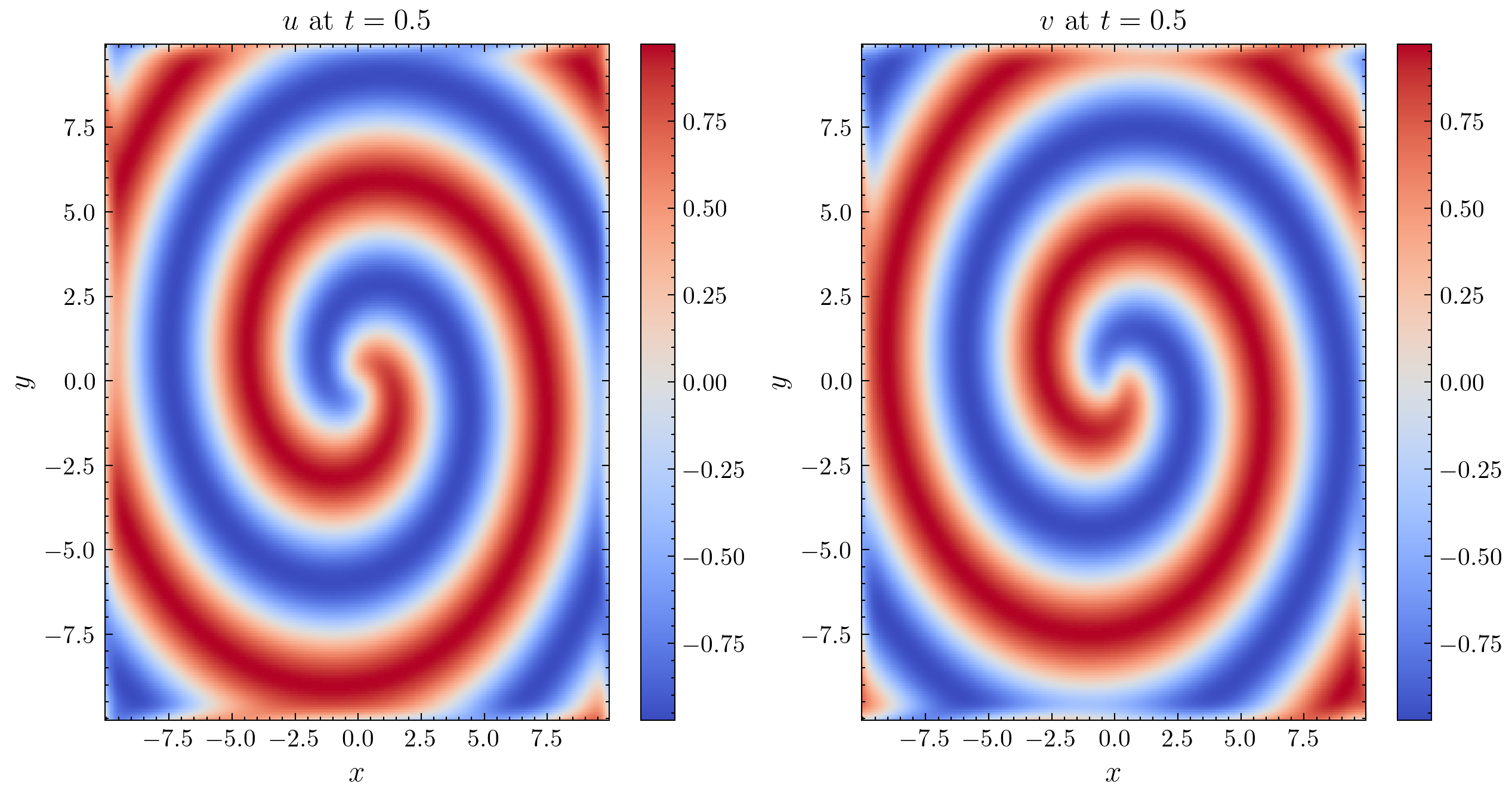}  
  \caption{RD}
\end{subfigure}
\begin{subfigure}{.49\textwidth}
  \centering
  \includegraphics[width=\linewidth]{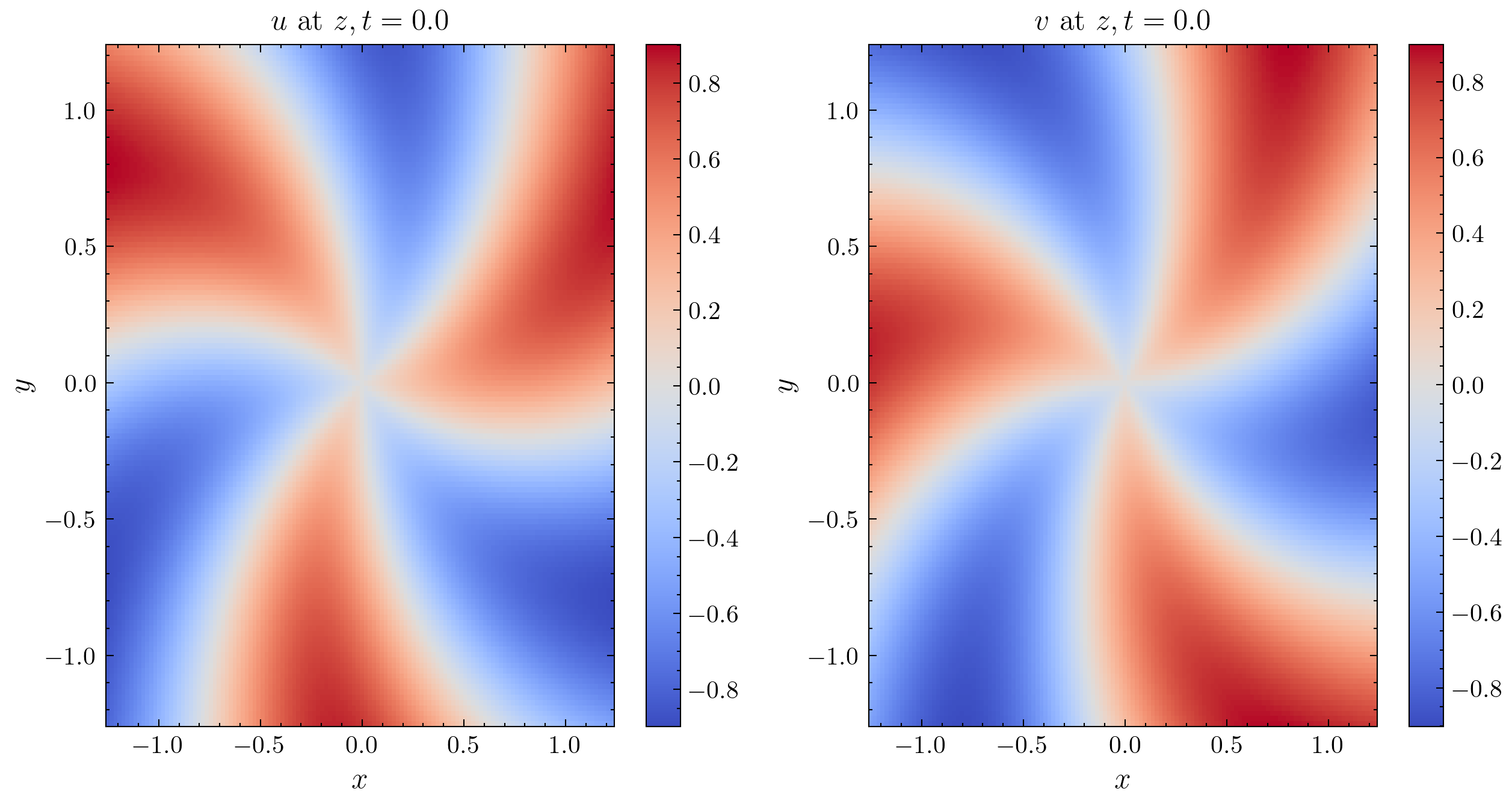}  
  \caption{GS}
\end{subfigure}
\caption{Dataset visualization}
\label{fig:data_vis}
\end{figure*}

\subsection{Data and Code Availability}
Please find the accompanying data and code made available at \url{https://github.com/Pongpisit-Thanasutives/UBIC}. We provide 2D visualization of the dataset files we experimented with in Figure \ref{fig:data_vis}. 
\end{document}